\documentclass[11pt]{article}

\usepackage[utf8]{inputenc}
\usepackage[T1]{fontenc}
\usepackage{lmodern}
\usepackage[margin=1in]{geometry}
\usepackage{microtype}
\usepackage{amsmath,amssymb}
\usepackage{booktabs}
\usepackage{array}
\usepackage{multirow}
\usepackage{graphicx}
\usepackage{xcolor}
\usepackage{enumitem}
\usepackage{caption}
\usepackage{subcaption}
\usepackage{tikz}
\usepackage{pgfplots}
\pgfplotsset{compat=1.18}
\usepackage{pgfplotstable}
\usepackage{algorithm}
\usepackage{algpseudocodex}
\usepackage{longtable}
\usepackage{tabularx}
\usepackage{tcolorbox}
\usepackage[numbers,sort&compress]{natbib}
\usepackage[hidelinks]{hyperref}
\usepackage{xurl}   
\usepackage{cleveref}
\usepackage{orcidlink}   

\definecolor{c0}{HTML}{0072B2}
\definecolor{c1}{HTML}{D55E00}
\definecolor{c2}{HTML}{009E73}
\definecolor{c3}{HTML}{E69F00}
\definecolor{c4}{HTML}{CC79A7}
\definecolor{cgray}{HTML}{888888}
\pgfplotsset{
  every axis/.append style={
    font=\small,
    line width=0.8pt,
    tick align=outside,
    grid=major,
    grid style={gray!20},
  },
}
\usetikzlibrary{patterns}

\newtcolorbox{cotbox}[1][]{colback=gray!4,colframe=gray!40,boxrule=0.5pt,
  fonttitle=\bfseries,fontupper=\ttfamily\footnotesize,#1}

\newcommand{\boxedans}{\texttt{\textbackslash boxed\{\}}}

\title{\textbf{A Verifiable Search Is Not a Learnable Chain-of-Thought}}

\author{%
  Harsh Patel\,\orcidlink{0009-0005-1704-811X}\\[2pt]
  \normalsize Independent Researcher\\[1pt]
  \normalsize \href{mailto:harshpatel2898@gmail.com}{harshpatel2898@gmail.com}
}
\date{June 2026}

\begin{document}
\maketitle

\begin{abstract}
It is tempting to assume that any task one can solve with a short program can be taught to a
model as its chain-of-thought: write the program's steps out in words, fine-tune on them, and
the model should follow. This paper shows the assumption fails for an identifiable class of
procedures, and that the reflexive remedies do not rescue it. The testbed is nine reasoning
tasks, each emitted by a deterministic generator, with one convenient property: the public and
hidden splits are drawn from the \emph{same} generators, so a held-out slice of training data
is a faithful proxy for test accuracy. I reverse-engineer the generators into Python solvers
(five reach $\geq\!98\%$), render their procedures as chain-of-thought, and distill them into a
rank-$\leq\!32$ LoRA over a 30B (3.5B-active) Nemotron model. Forward-computable tasks install
readily: four lookup/arithmetic tasks and an 8-bit boolean-rule task transfer ($\geq\!0.99$ and
$0.68$). Cryptarithm does not, when one tries to distill its backtracking \emph{search}: it holds
at $0.01$--$0.07$ across eleven chain-of-thought designs, RL from verifiable rewards, and
self-training, even though a search solver answers $71\%$ of its instances. The failure is not a
capability gap. The model does the arithmetic correctly on $97$--$100\%$ of lines and ranks the
correct cipher into its top eight on $71\%$ of instances. What it cannot do is carry the
\emph{search} forward as a left-to-right derivation. Supervised fine-tuning therefore learns the
\emph{shape} of a verifiable elimination step while its verdicts become unconditional templates,
correct only $16$--$57\%$ of the time (``verdict-as-token''). This search-distillation ceiling
holds across four backbones from 3B to 671B and across both fine-tuning and in-context prompting,
and a controlled intervention isolates the cause: revealing the cipher key, which turns the
derivation forward, lifts the \emph{same} instances from $0.03$ to $0.57$. The constraint binds
the \emph{search}, not the task: when a procedure's only solution is search over information-free
structure, no faithful forward chain-of-thought exists to imitate. The task nonetheless becomes
learnable by \emph{removing the search from the trace}, precomputing its combinatorial core into a
finite catalog the model memorizes and reducing the chain-of-thought to recall plus bounded
verification; the competition's 1st-place solution is the existence proof, reaching Private LB
$0.92$ by memorizing a per-signature candidate catalog rather than teaching the model to search
\citep{goodmeatday2026first}. What distills is memorization and verification, not search. I
release the benchmark decomposition, the solvers, the per-line fidelity audits, and the control
experiments.
\end{abstract}

\section{Introduction}
\label{sec:intro}

If you can write a short program that solves every instance of a task, is the task
``learnable'' as a chain-of-thought? Intuitively the program is a recipe, the recipe can
be written out as natural-language steps, and a capable model fine-tuned on those steps
should reproduce them. This intuition underlies a large body of recent work on distilling
reasoning into smaller models, from rationale distillation \citep{hsieh2023distill} to
bootstrapped self-training \citep{zelikman2022star} and reinforcement learning from
verifiable rewards \citep{shao2024deepseekmath, guo2025deepseekr1, lambert2024tulu3}.

This paper reports a controlled study in which the intuition fails, sharply and
reproducibly, for a specific and identifiable class of procedures. I use a benchmark of
nine reasoning tasks, each emitted by a deterministic generator (Roman numerals, linear
unit conversion, free-fall, monoalphabetic substitution cipher, an 8-bit boolean program,
two equation-induction families, and two cryptarithm families). The benchmark has a
property that makes it an unusually clean laboratory: the hidden test set is produced by
the \emph{same} generators as the public training set, so the two are draws from one
distribution and a stratified held-out slice of training data predicts test accuracy
directly. I can therefore measure ``did the procedure transfer'' to a fraction of a
point without access to the test labels.

\paragraph{Research questions.} I ask: \textbf{(RQ1)} does a verifiable solver's coverage
predict whether its procedure is learnable as a small model's chain-of-thought?
\textbf{(RQ2)} when a procedure resists every training method I try, what is the binding
bottleneck? \textbf{(RQ3)} is there a cheap, pre-training test that predicts resistance? In
short: \emph{no}, solver coverage does not predict whether the \emph{search} distills
(\Cref{fig:thesis}); a search over information-free structure admits no faithful forward CoT, so
SFT learns only its unfaithful surface form (``verdict-as-token''), architecture-independently
(\Cref{sec:res-crypt}). The binding bottleneck is forward-derivability, isolated by a controlled
intervention that lifts the same instances $0.03\!\to\!0.57$ (\Cref{tab:interv}). The resistance
is predictable in advance, via a three-part screening heuristic (\Cref{sec:analysis}). And it is
not a ceiling on the \emph{task}: the practical escape, demonstrated by the competition's
1st-place solution, is to \emph{memorize} the search's finite structure rather than distill the
search itself (\Cref{sec:memo}).

I first reverse-engineer the generators into Python solvers. Five of the nine tasks admit
solvers at $\geq\!98\%$ accuracy. Cryptarithm admits a backtracking-search solver that
answers $\approx\!71\%$ of instances (the remainder are information-theoretically
ambiguous). I then attempt to install each solver's procedure into a rank-$\leq\!32$
LoRA adapter \citep{hu2022lora} over \texttt{Nemotron-3-Nano-30B-A3B}, a hybrid
Mamba-2/Mixture-of-Experts model \citep{nemotron3nano2025, dao2024mamba2}, served greedily
under a 7680-token budget. Training is SFT on solver-grounded synthetic CoT generated with
\emph{different} values from any evaluation instance (zero leakage); when SFT plateaus I
escalate to RLVR/GRPO and STaR.

\paragraph{Thesis.} A verifiable \emph{search} does not distill into a chain-of-thought. The
dividing line is whether the task admits a \emph{faithful forward chain-of-thought}, a
left-to-right derivation that reaches the answer. Forward-computable tasks install readily; a task
whose only solution is backtracking search over information-free structure has no faithful forward
trace, so SFT copies the \emph{shape} of the search but not its logic, across every architecture
and scale I tried. The task is not thereby unlearnable: it becomes learnable by \emph{removing the
search from the trace}, precomputing its combinatorial core into a finite catalog the model
memorizes and reducing the chain-of-thought to recall plus bounded verification (\Cref{sec:memo}).
What distills is memorization and verification, not search.

\paragraph{Contributions.}
\begin{itemize}[leftmargin=1.4em,itemsep=2pt,topsep=2pt]
\item \textbf{A controlled solvability--learnability gap.} On a train$\equiv$test
  benchmark I quantify, per task, the gap between a verifiable solver's accuracy and the
  accuracy the fine-tuned model actually reproduces in CoT (\Cref{fig:thesis},
  \Cref{tab:master}). The gap is near-zero for four tasks and exceeds $0.65$ for
  cryptarithm.
\item \textbf{A negative result with convergent evidence.} Across eleven CoT-engineering
  rounds and three RL/self-training escalations, cryptarithm-deduce never exceeds $0.07$
  (\Cref{fig:crypt}). Five independent measurements, solver ceiling, information floor,
  base-model behavior, forward-derivation coverage, and per-line execution fidelity, each
  independently bound the achievable accuracy, and they agree (\Cref{tab:walls}).
\item \textbf{Three named, measured mechanisms.} \emph{Verdict-as-token}
  (\Cref{fig:fidelity}): arithmetic is correct on $97$--$100\%$ of lines but elimination
  \emph{verdicts} are emitted as unconditional templates (fidelity $16$--$57\%$).
  \emph{Derivation- vs.\ ranking-blocked}: a budget-unbounded generate-and-verify reframe
  shows ranking is not the constraint ($\mathrm{hit@}8=0.71$) while forward derivation is
  ($1/659$). \emph{Information-free sub-structure}: the hidden map's
  $\mathrm{MI}(\text{digit};\text{glyph})=0.021$ nats $\approx 0$.
\item \textbf{A positive control that confirms the mechanism.} Hand-written bit-manipulation
  CoT that \emph{teleports} the rule search collapses to $\approx\!0.05$; STaR on the
  model's own successful searches lifts $0.526\!\to\!0.678$ and removes truncation
  ($18.6\%\!\to\!0.2\%$, \Cref{fig:bitstar}). What transfers is search the model can
  actually run, not search narrated for it.
\item \textbf{An architecture control.} Training the identical cryptarithm corpus on four
  backbones, hybrid Mamba/MoE, two dense Transformers, and an MoE Transformer, spanning
  $3$B--$21$B, hits the \emph{same} $\leq\!0.04$ floor (\Cref{tab:archctrl}). The
  search-distillation ceiling is architecture-general, not backbone- or scale-specific. This
  refutes my own initial SSM-specific hypothesis.
\item \textbf{A causal intervention.} Revealing the cipher key on the \emph{same} cryptarithm
  instances, turning the derivation forward, lifts Nemotron from $0.03$ to $\mathbf{0.571}$
  (\Cref{tab:interv}). Revealing only half barely helps. Forward-derivability is the causal
  lever, and a faithful forward CoT must cover the \emph{whole} derivation, not most of it.
\item \textbf{The escape, externally validated.} The search-distillation ceiling is \emph{not} a
  ceiling on the task. The competition's 1st-place solution reaches Private LB $0.92$ by
  \emph{memorizing} the search's finite structure, a $4{,}205$-entry per-signature candidate
  catalog for cryptarithm and a $5{,}238$-entry rule-sequence catalog for bit-manipulation, and
  verifying candidates in-trace, rather than distilling the search \citep{goodmeatday2026first}.
  This both confirms the mechanism, why memorization is necessary, and bounds the claim
  (\Cref{sec:memo}).
\end{itemize}

I did not win the competition this benchmark is drawn from, my best adapter scored
$0.85$ public ($0.86$ private, on an unselected submission, \Cref{sec:dynamics}) against a
leader at $0.92$. As I discuss in \Cref{sec:dynamics}, $0.85$ is the
\emph{median} of the $4{,}355$-team field and the ceiling reachable without the
memorization--computation reformulation of cryptarithm and bit-manipulation (\Cref{sec:memo});
a fully open-source solution taking my same search-distillation approach won the competition's
Open Progress Prize at the same $0.85$. The contribution is not a ranking; it is
a clean, instrumented account of \emph{where and why} a verifiable search fails to become
model reasoning, and of what does transfer in its place.

\section{Background and Related Work}
\label{sec:related}

\paragraph{Chain-of-thought and its distillation.} Eliciting step-by-step reasoning
improves LLM accuracy on multi-step tasks \citep{wei2022cot, wang2023selfconsistency}, and
a model's rationales can be distilled into smaller students
\citep{hinton2015distill, hsieh2023distill, muennighoff2025s1}. My setting is distillation
in the strong sense: the ``teacher'' is a perfect symbolic solver, and I ask whether its
procedure survives transfer into a small model's CoT under a budget.

\paragraph{Self-training and verifiable rewards.} STaR bootstraps a model on its own
correct rationales \citep{zelikman2022star}; RL from verifiable rewards and group-relative
policy optimization (GRPO) optimize directly against an answer checker
\citep{shao2024deepseekmath, guo2025deepseekr1, lambert2024tulu3}, and process verifiers
supervise intermediate steps \citep{lightman2023verify}. The generate-and-verify idea traces
to training answer verifiers on grade-school math \citep{cobbe2021gsm8k}. I use all three as
escalations. A central finding is that on my hard task RLVR's gradient is \emph{present but
non-transferring}: it improves easy curricula yet does not move held-out accuracy, because
positive rollouts on real instances are too rare to learn from.

\paragraph{Faithfulness of chain-of-thought.} A growing literature shows that a model's
stated reasoning need not reflect the computation behind its answer: explanations can be
systematically unfaithful \citep{turpin2023unfaithful}, and interventions on the trace
(inserting mistakes, truncating, paraphrasing) often leave the answer unchanged
\citep{lanham2023measuring}, which motivates constructions that force the answer to follow
from the stated steps \citep{lyu2023faithful}. My \emph{verdict-as-token} result is a sharp,
mechanistic instance of unfaithfulness arising in the \emph{training} regime: SFT installs the
surface form of a verifiable elimination step while decoupling it from the step's logic, so
the imitated trace is unfaithful by construction at free-running inference.

\paragraph{Limits of transformers on procedures.} My negative result is consistent with,
and sharpens, work on compositional and algorithmic limits: transformers degrade on
multi-step composition as depth grows \citep{dziri2023faith} and struggle to generalize
algorithm length \citep{anil2022length}. Length generalization holds only for tasks
expressible as short length-invariant programs \citep{zhou2024algorithms}, memory-free
architectures cannot climb the Chomsky hierarchy \citep{deletang2023chomsky}, algorithmic
competence can emerge only after delayed generalization \citep{power2022grokking}, and models
score poorly even when worked step-by-step solutions exist \citep{hendrycks2021math}. I add a
\emph{mechanism} specific to imitation learning of search, verdict-as-token under teacher
forcing, and connect it to the classic
exposure-bias gap between teacher-forced training and free-running inference
\citep{bengio2015scheduled, ranzato2016sequence}.

\paragraph{Architecture.} The base model is a hybrid in which most attention layers are
replaced by Mamba-2 state-space layers, interleaved with a sparse Mixture-of-Experts
\citep{nemotron3nano2025, dao2024mamba2}, and served with paged-attention batching
\citep{kwon2023vllm}. State-space layers compress history into a fixed-size state; I will
argue this interacts unfavorably with tasks that require explicit, unbounded search-state.
Only $6$ of the model's $\sim\!52$ layers carry attention; the bulk are Mamba-2 or MoE. The
rank-$\leq\!32$ adapter I train spans the attention, Mamba, MoE, and output projections
alike (\Cref{app:training}), but a rank-$32$ update is a thin slice of a model whose capacity
sits overwhelmingly in layers that summarize history into fixed state. There is theory for why
this should bite: state-space layers lie in $\mathrm{TC}^0$ and cannot perform genuine
sequential state-tracking \citep{merrill2024illusion}, and a fixed-size recurrent state caps
copying and retrieval from context \citep{jelassi2024repeat}. I tested this directly, however,
and find it is \emph{not} the explanation for cryptarithm (\Cref{tab:archctrl}): dense and
MoE-Transformer models of comparable active size hit the same floor, so the SSM limit is at most
a contributing factor (e.g.\ for the bit-manipulation tail), not the cause of the search-distillation ceiling.

\section{The Testbed: a Deterministic-Generator Benchmark}
\label{sec:testbed}

\paragraph{Task and grading.} The benchmark ships a public training set of $9{,}500$
labeled problems and is scored on a hidden test set of $\approx\!500$ problems. A
submission is a single LoRA adapter of rank at most $32$ loaded onto the frozen base model
and served with vLLM in deterministic (greedy) mode (temperature $0$, top-$p$ $1.0$,
\texttt{max\_tokens}$=7680$, \texttt{max\_model\_len}$=8192$), with thinking enabled
(\verb|<think>...</think>|). The model must place its final answer in a \boxedans{} field;
the grader extracts the last such field (falling back to the last number) and applies a
fixed \texttt{verify()}: exact case-insensitive string match for binary strings,
relative tolerance $10^{-2}$ for floats, exact string match otherwise.

\paragraph{The train$\equiv$test property.} Every category is produced by a deterministic
generator, and the hidden test uses the \emph{same} generators. Train and test are
therefore i.i.d.\ draws from one distribution, and a stratified held-out slice of
\texttt{train.csv} (100 rows/category) is a faithful oracle for the leaderboard. I
exploit this throughout: I train on synthetic CoT built with different values
(zero leakage) and evaluate on real held-out training rows, treating that number as the
test accuracy. A determinism check confirms the regime: when an in-class rule fits a
problem's worked examples, it yields a unique prediction on the query $98.7\%$ of the time
and is correct $99.8\%$ of the time, the examples \emph{pin} the rule, so the only limits
to $100\%$ are solver and model completeness, not intrinsic ambiguity (cryptarithm, below,
is the exception that proves this).

\paragraph{The nine generators.} \Cref{tab:master} lists the categories, their test weights,
and the base model's zero-shot accuracy. Briefly: \emph{numeral} (write an integer in Roman
numerals), \emph{unit\_conversion} (a linear map $y=ax+b$ pinned by two examples),
\emph{gravity} (free-fall $d=\tfrac12 g t^2$, a one-parameter fit), and \emph{cipher}
(a monoalphabetic substitution over a fixed 77-word vocabulary, solved by aligning example
word-pairs and constraint-propagating the rest) are the ``easy four.''
\emph{bit\_manipulation} maps 8 bits to 8 bits by a single global boolean function over up
to three \emph{taps} (rotated/shifted/negated copies of the input), the same function for
all eight output bits, pinned by 7--10 examples. \emph{equation\_numeric} (deduce/guess)
shows visible-digit equations $AB\,[\text{op}]\,CD = \text{RHS}$ from which one must induce
the operand pairing, the operation, and an output-string format. \emph{cryptarithm}
(deduce/guess) is the hard case, described next.

\paragraph{Cryptarithm structure.} Each example is a five-character left-hand side
\texttt{s0 s1 OP s2 s3 = RHS}. Three things are hidden: (1) a per-row injective
digit$\to$symbol cipher (a uniform draw of 10 distinct glyphs from a 23-symbol pool);
(2) an \emph{operation} attached to the middle operator glyph, drawn from a wide vocabulary
(add, sub, mul, abs-difference, mod, gcd, lcm, floor-div, the $\pm1/\pm2$ variants, and
two purely structural \emph{concat} operations). And (3) a per-row \emph{endianness}
(standard or little-endian, reversing operand and result strings). ``Deduce'' means the
query's operator was seen in the examples; ``guess'' means it was not. Because the cipher
is information-free, no single column reveals a digit: the map is pinned only by requiring
\emph{all} example equations to hold simultaneously under the right operations and
direction, a cross-equation constraint search with backtracking. This is the structural
reason cryptarithm sits on the far side of the learnability frontier.

\paragraph{Compute and budget.} Final adapters are trained on a single RTX PRO 6000
(Blackwell, 96\,GB) via the competition platform. The adapter has $\approx\!880$M trainable
parameters ($2.71\%$ of the model). Cheap iteration on data and CoT design is done on a
managed LoRA service at $\approx\!\$3.5$/round before promoting a finalized dataset to the
full training run, a three-tier ``cheapest-first'' ladder (API probe $\to$ cheap LoRA
$\to$ full train) that I found essential for keeping a long negative-result search
affordable. All CoT is written in plain ASCII: the tokenizer is a $131{,}072$-entry BPE
with no \texttt{<unk>}, but Unicode operators ($\oplus,\neg,\neq$) fragment into rare
multi-byte tokens, so I spell operators as words (\texttt{XOR}, \texttt{NOT}),
arrows as \verb|->|, and inequality as \verb|!=|.

\begin{table}[tbp]
\centering
\caption{The benchmark, and the central gap. ``Solver'' is the accuracy of my
reverse-engineered Python program (the data-generation oracle); ``Model'' is the accuracy
the fine-tuned rank-$\leq\!32$ adapter actually reproduces in greedy CoT, measured on a
held-out slice of real training rows (the faithful test oracle). The gap is near-zero for
the lookup/arithmetic tasks and enormous for cryptarithm, despite a $0.71$ search-solver.}
\label{tab:master}
\small
\begin{tabular}{l r r r r}
\toprule
Category & Test wt. & Base & \textbf{Solver} & \textbf{Model} \\
\midrule
numeral             & 16.6\% & 0.959 & 1.00 & $\approx$1.00 \\
unit\_conversion    & 16.8\% & 0.745 & 1.00 & $\approx$1.00 \\
gravity             & 16.8\% & 0.585 & 1.00 & $\approx$1.00 \\
cipher              & 16.6\% & 0.306 & 1.00 & $\approx$0.99 \\
bit\_manipulation   & 16.9\% & 0.080 & 0.989 & \textbf{0.678} \\
equation\_num.\ (deduce) & 6.3\% & 0.287 & 0.951 & 0.83 \\
equation\_num.\ (guess)  & 1.4\% & 0.037 & 0.206$^\dagger$ & 0.18 \\
cryptarithm (deduce)& 6.9\% & 0.003 & 0.71$^\ddagger$ & \textbf{0.01--0.07} \\
cryptarithm (guess) & 1.7\% & 0.000 & 0.59$^\ddagger$ & 0.02 \\
\midrule
\multicolumn{2}{l}{\emph{Overall (best banked adapter)}} & 0.466 &, & \textbf{0.85} \\
\bottomrule
\end{tabular}
\\[2pt]
{\footnotesize $^\dagger$ The operation is information-free given the symbol, so $0.206$ is
the marginal-prior ceiling, not a coverage limit. $^\ddagger$ Search-based solver
(backtracking / generate-and-verify). The \emph{forward-derivable} fraction, what a
left-to-right CoT could honestly imitate, is $\approx\!0$ (\Cref{sec:res-crypt}). Model
$95\%$ Wilson CIs ($n$): bit $[.64,.72]$ ($500$); eq-deduce wide ($n{=}18$, see
\Cref{app:failstats}); eq-guess $[.12,.28]$ ($96$); crypt-deduce $[.01,.09]$ ($93$);
crypt-guess $[.01,.08]$ ($87$); easy-four/cipher $n\!\approx\!1.6$k (CI $\le\!\pm0.4$pp).}
\end{table}

\begin{figure}[tbp]
\centering
\begin{tikzpicture}
\begin{axis}[
  width=0.92\textwidth, height=6.0cm,
  ybar, bar width=5pt,
  ymin=0, ymax=1.08,
  ylabel={Accuracy},
  symbolic x coords={num/unit/grav, cipher, bit\_manip, eq\_deduce, eq\_guess, crypt\_deduce, crypt\_guess},
  xtick=data, x tick label style={rotate=20,anchor=east,font=\footnotesize},
  ytick={0,0.2,0.4,0.6,0.8,1.0},
  legend style={at={(0.5,1.03)},anchor=south,legend columns=3,draw=none,font=\footnotesize},
  enlarge x limits=0.08,
]
\addplot[fill=c0,draw=c0] coordinates {
  (num/unit/grav,1.0) (cipher,1.0) (bit\_manip,0.989) (eq\_deduce,0.951)
  (eq\_guess,0.206) (crypt\_deduce,0.71) (crypt\_guess,0.59)};
\addplot[fill=c3,draw=c3] coordinates {
  (num/unit/grav,1.0) (cipher,1.0) (bit\_manip,0.95) (eq\_deduce,0.95)
  (eq\_guess,0.206) (crypt\_deduce,0.10) (crypt\_guess,0.05)};
\addplot[fill=c1,draw=c1] coordinates {
  (num/unit/grav,1.0) (cipher,0.99) (bit\_manip,0.678) (eq\_deduce,0.83)
  (eq\_guess,0.18) (crypt\_deduce,0.05) (crypt\_guess,0.02)};
\legend{Solver (search), Forward-derivable, Model (greedy)}
\end{axis}
\end{tikzpicture}
\caption{\textbf{The learnability frontier.} Solver (backtracking search) vs.\ the
\emph{forward-derivable} ceiling (what a left-to-right CoT could honestly imitate) vs.\ the
fine-tuned model. For lookup/fit tasks all three coincide. For cryptarithm the solver reaches
$0.71$, but the forward-derivable ceiling collapses to $\approx\!0.10$ and the model tracks
\emph{that}, not the solver ($0.05$). bit\_manipulation is the exception where forward
derivation is high yet the model still gaps, a search-depth limit STaR partly closes
(\Cref{sec:res-easy}). The ``guess'' tasks are information-limited. Exact values and CIs in
\Cref{tab:master}.}
\label{fig:thesis}
\end{figure}
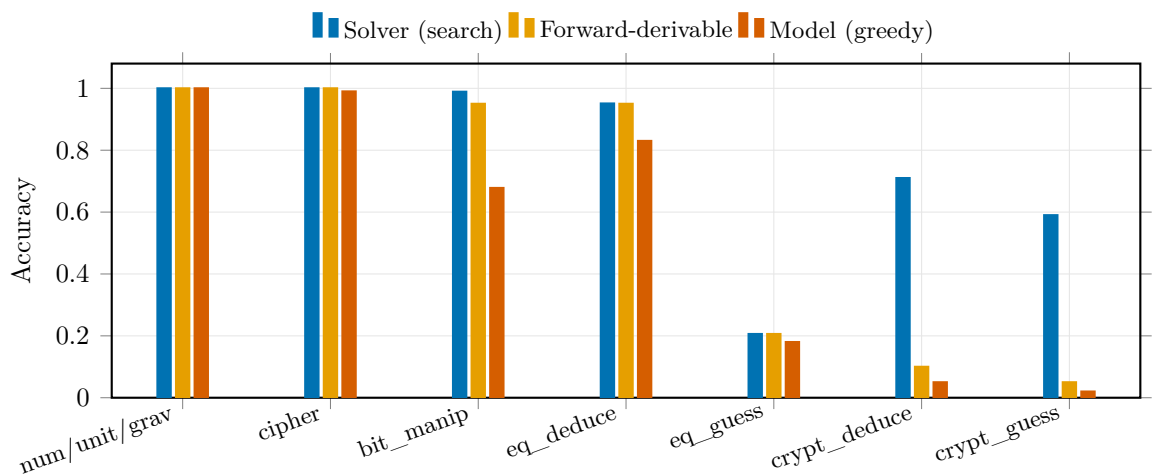

\section{Method: Solver-Grounded Synthetic CoT and the Experiment Ladder}
\label{sec:method}

My pipeline is deliberately simple, so that any failure is attributable to learnability
rather than to data noise. For each category I (1) reverse-engineer the generator into a
solver and validate it against the $9{,}500$ training rows; (2) \emph{render} a synthetic
CoT for fresh, randomly generated instances by instrumenting the solver to narrate its own
steps in the base model's native ASCII style, ending in a \boxedans{}; (3) SFT the LoRA on
this synthetic corpus. And (4) evaluate the adapter on held-out \emph{real} training rows.
Because step (2) uses different values than any evaluation row, there is no leakage; because
of the train$\equiv$test property, step (4) is the leaderboard.

I deliberately enforce one CoT discipline throughout: the trace must \emph{show the search}
(try $\to$ reject $\to$ match) and must never teleport to an answer it has not derived on
the page. A trace that states the conclusion without the work teaches nothing transferable;
this principle, and its violations, turn out to be the crux of the whole study
(\Cref{sec:analysis}).

\paragraph{Rendering, and what ``witnessed'' means.} I render a CoT by instrumenting the
solver to narrate each step in the base's native ASCII voice: transcribe the relevant glyphs,
compute, and emit a \emph{verdict} (``keep,'' ``drop,'' ``this run extends''). The one
non-negotiable rule is that every verdict must \emph{restate, on its own line, the evidence
that forces it}, a \emph{witnessed} verdict. A verdict that is a fixed phrase decoupled from
the line's own numbers is a \emph{teleport}: it reads as reasoning but carries none.
\Cref{fig:cot} contrasts the two on a real ones-digit elimination; the failure shown is
verbatim from a trained model and is the modal cryptarithm error (\Cref{fig:fidelity}).

\begin{figure}[htbp]
\centering
\begin{cotbox}[title=\normalfont\bfseries Witnessed verdict (rendered) vs.\ verdict-as-token (emitted)]
WITNESSED:\ EQ1 ones: target ones = 4 (= b ones). try a ones 6: 6*4 ends 4 -> MATCH -> keep 6.\\[2pt]
TELEPORT:\ \ EQ1 ones: ... 3*4 ends 2; 5*4 ends 0; 6*4 ends 4; 7*4 ends 8; 8*4 ends 2; 9*4 ends 6 -> none matches -> drop.
\end{cotbox}
\caption{\textbf{Witnessed vs.\ teleported verdict.} The teleport line is verbatim from a
trained model's transcript (\texttt{10b71e8a}): it computes ``\texttt{6*4 ends 4}'', which
\emph{matches} the target ones digit $4$, on the very same line, then concludes ``none matches
$\to$ drop,'' wrongly eliminating the correct operation. The arithmetic is correct; the verdict
does not follow from it. SFT on traces whose verdicts are not witnessed installs exactly this
behavior.}
\label{fig:cot}
\end{figure}

When SFT plateaus on a category I escalate along two axes: RLVR/GRPO against the binary
\texttt{verify()} reward, and STaR, harvesting the model's \emph{own} correct rollouts
(filtered by the verifier) and folding them back as SFT data. The escalations cost more, so
I gate them: a round is only promoted if a cheap held-out probe clears a pre-registered
threshold.

\paragraph{Statistical reporting.} Accuracies are point estimates reported with Wilson
$95\%$ confidence intervals and the evaluation size $n$. Held-out sets are real-train slices
(\Cref{sec:testbed}): $100$/category unless noted, with bit-manipulation evaluated on a
disjoint $500$-row set and cryptarithm on the $n{=}93$/$87$ ``unseen'' deduce/guess slices.
I flag explicitly where small $n$ makes a difference non-significant (it does for the
$\mathrm{pass@}k$ comparisons. It does \emph{not} for the headline solver--model gaps).

\section{Results}
\label{sec:results}

My version-by-version leaderboard trajectory (\Cref{app:lbtraj}) situates the project. The
frozen base model scores $0.466$; a null
(zero) adapter reproduces it ($0.50$). Installing the four easy tasks and a first
bit-manipulation CoT reaches $0.59$. A from-base adapter with finalized
cipher/bit/equation grammars banks $0.85$. On the final $4{,}355$-team leaderboard $0.85$ is
the \emph{median}: $2{,}236$ teams reached $\geq\!0.85$, only $66$ cleared $0.87$, and just
the leader reached $0.92$ (\Cref{sec:dynamics}). My $0.85$ is the solver ceiling of the
tasks I could make learnable; the missing $0.07$ to the top is almost entirely cryptarithm
and the bit-manipulation tail. The rest of this section dissects those two.


\subsection{Where the procedure transfers: the easy four and bit\_manipulation}
\label{sec:res-easy}

The four lookup/fit tasks are uninteresting in the best way: once the CoT is correct and
ASCII-clean, the model reproduces the solver to $\geq\!0.99$. They establish that the
pipeline, the budget, and the grader are not the bottleneck.

\textbf{Bit-manipulation} is the informative middle of the frontier, and the place where
knowing the generator pays off most. The task's own prompt names its operations:
\emph{``The transformation involves operations like bit shifts, rotations, XOR, AND, OR, NOT,
and possibly majority or choice functions.''} This is a gift, it says the hidden rule is a
single boolean function over up to three \emph{taps} (rotated/shifted copies of the 8-bit
input), drawn from a small human-nameable vocabulary and applied identically to all eight
output bits. I make that explicit: a six-function basis $\{\textsc{xor}, \textsc{maj},
\textsc{xor\_or}, \textsc{or}, \textsc{or\_xnor}, \textsc{choice}\}$ (two 2-argument and four
3-argument functions; \Cref{tab:basis}) accounts, under a first-match partition, for $1577$
of the $1602$ training rows. The remaining $25$ need a deeper composition. \textsc{and}-of-three
is redundant, its rows are subsumed by majority ($108/109$). This is the same single-rule
space my global solver searches by op-composition to $98.9\%$ query accuracy
(\Cref{app:bit}). The basis is its interpretable face, mapping one-to-one onto the
generator's stated vocabulary.

The point is the contrast. The \emph{solver} side of bit-manipulation is essentially closed:
a handful of named functions over a handful of taps explains the entire dataset. Against
that, the fine-tuned model's $0.678$ is not a data problem, it is a learnability problem,
and a sharply structured one (\Cref{tab:bittaps}): the rules the model misses are
overwhelmingly the three-tap ones ($34\%$ of the category, accuracy $0.50$), exactly the
rules whose search is deepest. The procedure, enumerate the basis $\times$ tap-assignments,
verify on all examples, apply, is the bounded search I expected to be teachable. It is, but
only in one specific way.

\begin{table}[htbp]
\centering
\caption{A six-function basis for the bit-manipulation rule, mapping onto the generator's own
prompt vocabulary. Each function takes up to three \emph{taps} (independent rotated/shifted
copies of the input) and is applied identically to all eight output bits. Counts are a
priority-ordered first-match partition of the $1602$ training rows.}
\label{tab:basis}
\small
\begin{tabular}{l c l r}
\toprule
Function & Arity & Definition over taps $(a,b,c)$ & Rows (first-match) \\
\midrule
\textsc{xor}      & 2 & $a\oplus b$ & 625 \\
\textsc{maj}      & 3 & $(a\wedge b)\vee(a\wedge c)\vee(b\wedge c)$ & 459 \\
\textsc{xor\_or}  & 3 & $(a\oplus b)\vee c$ & 234 \\
\textsc{or}       & 2 & $a\vee b$ & 148 \\
\textsc{or\_xnor} & 3 & $a\vee\overline{(b\oplus c)}$ & 58 \\
\textsc{choice}   & 3 & $(a\wedge b)\vee(\bar a\wedge c)$ & 53 \\
\midrule
\multicolumn{3}{l}{\emph{covered by basis}} & \textbf{1577 / 1602} \\
\textsc{and3}     & 3 & $a\wedge b\wedge c$ & \emph{redundant} ($\subset\textsc{maj}$, 108/109) \\
\bottomrule
\end{tabular}
\\[2pt]
{\footnotesize The split is a first-match partition under a fixed priority order; an
independent re-derivation confirms \textsc{xor} ($\approx\!625$--$638$) and \textsc{maj}
($\approx\!459$--$478$) as the two largest and robust strata, while the exact division within
the \{\textsc{or},\textsc{xor\_or},\textsc{or\_xnor},\textsc{choice}\} cluster depends on the
ordering (its total does not). The op-composition solver attains $98.9\%$ query accuracy on
the same rows (\Cref{app:bit}).}
\end{table}

Every \emph{hand-written} CoT I authored collapsed to $\approx\!0.05$--$0.06$: a per-bit
``dialect'' (0.046), a hand grammar that named the taps then asserted the rule (0.062), and
a peel-to-two-tap scheme (0.053). The common defect, confirmed by a line-level autopsy of 28
wrong output bits (\Cref{sec:analysis}), is that each authored trace \emph{teleports} the
one step that matters, the rule search, because a human author already knows the answer
and cannot un-know it while writing. Transcription, arithmetic, and assembly in these traces
are flawless. Only the search verdict is fake.

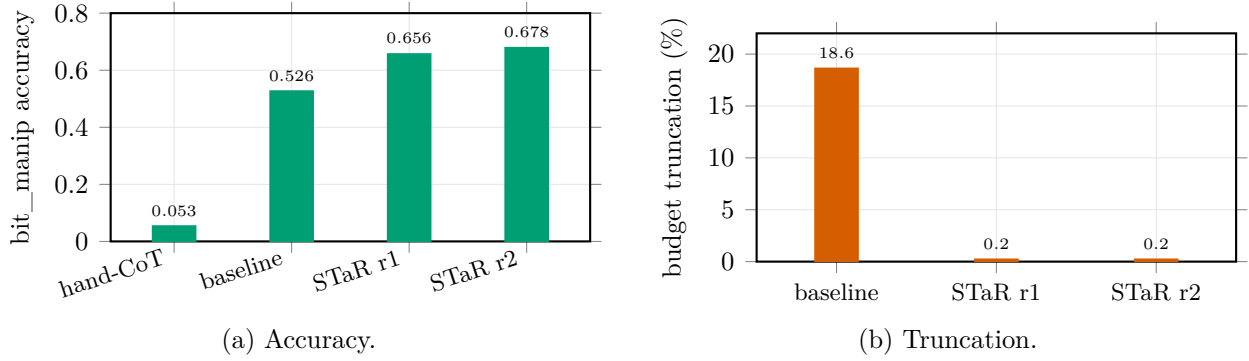
\begin{figure}[tbp]
\centering
\begin{subfigure}{0.48\textwidth}
\centering
\begin{tikzpicture}
\begin{axis}[
  width=\textwidth, height=4.6cm,
  ybar, bar width=16pt, ymin=0, ymax=0.8,
  ylabel={bit\_manip accuracy},
  symbolic x coords={hand-CoT, baseline, STaR r1, STaR r2},
  xtick=data, x tick label style={rotate=18,anchor=east,font=\footnotesize},
  nodes near coords, nodes near coords style={font=\tiny,/pgf/number format/fixed,/pgf/number format/precision=3},
  enlarge x limits=0.18,
]
\addplot[fill=c2,draw=c2] coordinates {
  (hand-CoT,0.053) (baseline,0.526) (STaR r1,0.656) (STaR r2,0.678)};
\end{axis}
\end{tikzpicture}
\caption{Accuracy.}
\end{subfigure}\hfill
\begin{subfigure}{0.48\textwidth}
\centering
\begin{tikzpicture}
\begin{axis}[
  width=\textwidth, height=4.6cm,
  ybar, bar width=16pt, ymin=0, ymax=22,
  ylabel={budget truncation (\%)},
  symbolic x coords={baseline, STaR r1, STaR r2},
  xtick=data, x tick label style={font=\footnotesize},
  nodes near coords, nodes near coords style={font=\tiny,/pgf/number format/fixed,/pgf/number format/precision=1},
  enlarge x limits=0.25,
]
\addplot[fill=c1,draw=c1] coordinates {
  (baseline,18.6) (STaR r1,0.2) (STaR r2,0.2)};
\end{axis}
\end{tikzpicture}
\caption{Truncation.}
\end{subfigure}
\caption{\textbf{Bit-manipulation: only the model's own search transfers.} Hand-written CoT
that teleports the rule search scores like the base model ($0.053$). STaR on
verifier-passed self-traces lifts accuracy and collapses budget truncation, because the
imitated traces are searches the model can actually run within budget. Baseline$\to$STaR is
significant (disjoint $95\%$ CIs, $n{=}500$). The r1$\to$r2 step is within noise.}
\label{fig:bitstar}
\end{figure}

What worked was STaR. Seeding from a weak warm start, I harvested the model's \emph{own}
verifier-passed rollouts and fine-tuned on them. Accuracy rose $0.526\!\to\!0.656\!\to\!0.678$
over two rounds, and, tellingly, budget truncation fell from $18.6\%$ to $0.2\%$
(\Cref{fig:bitstar}). The model's own successful searches are short, decisive, and (because
they actually ran) honest. Imitating them installs a procedure the model can execute at
inference, closing the exposure-bias gap that the hand-written traces opened. Accuracy then
plateaus: the residual $\approx\!0.50$ on genuine three-tap rules is the base's search
ceiling on this architecture, and additional synthetic three-tap data did not beat the STaR
number ($0.602 < 0.678$).

\subsection{Where it does not: cryptarithm}
\label{sec:res-crypt}

\begin{figure}[tbp]
\centering
\begin{tikzpicture}
\begin{axis}[
  width=0.92\textwidth, height=5.2cm,
  ybar, bar width=12pt, ymin=0, ymax=0.78,
  ylabel={cryptarithm\_deduce accuracy},
  symbolic x coords={r1,r2,r3,r5,r6,r7,r8,r9,r10},
  xtick=data, x tick label style={font=\footnotesize},
  ytick={0,0.2,0.4,0.6},
  nodes near coords, nodes near coords style={font=\tiny,/pgf/number format/fixed,/pgf/number format/precision=2},
  enlarge x limits=0.07,
]
\addplot[fill=c1,draw=c1] coordinates {
  (r1,0.03) (r2,0.07) (r3,0.05) (r5,0.04) (r6,0.05) (r7,0.07) (r8,0.05) (r9,0.01) (r10,0.01)};
\draw[c0,thick,dashed] (axis cs:r1,0.71) -- (axis cs:r10,0.71)
  node[above right,font=\footnotesize,c0,xshift=-30pt] {search-solver $\approx$0.71};
\draw[c2,thick,dotted] (axis cs:r1,0.10) -- (axis cs:r10,0.10)
  node[above,font=\footnotesize,c2] {forward-derivable $\lesssim$0.10};
\end{axis}
\end{tikzpicture}
\caption{\textbf{One floor, many rounds.} Cryptarithm-deduce accuracy under successive CoT
designs (r1--r11. The nine reaching a held-out greedy eval are shown; RLVR/GRPO and the
generate-and-verify reframe ran between rounds; \Cref{tab:ledger} lists all fourteen training
runs). The
search-solver answers $\approx\!71\%$ of instances. The fraction a faithful forward CoT
could imitate is $\lesssim\!10\%$; the model lands at $\approx\!0.05$ regardless of design
effort.}
\label{fig:crypt}
\end{figure}
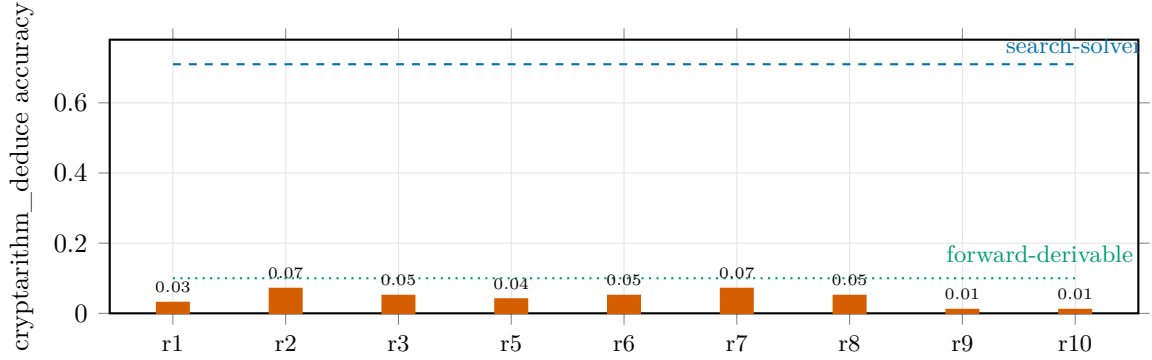

Cryptarithm is the negative result, and it is robust. I ran eleven successive CoT designs
(r1--r11; \Cref{fig:crypt} plots the nine that reached a held-out greedy eval, r4 having been
gate-failed before training): decode-first verification, truthful
conditionals, policy-driven backtracking renderers, two-regime structure modeling,
construction-of-symbol-table scaffolds, monotone try-counters with bail-out endings,
self-routing forced chains, witness-bearing elimination lines, a generate-and-verify
reframe, and explicit terminate-on-exhaustion traces. Every round lands in $[0.01,0.07]$.
The truncation rate swings wildly across rounds, from $87\%$ down to $1\%$, while accuracy
does not move, which already rules out ``ran out of budget'' as the explanation.

\paragraph{Five walls converge.} I then measured the ceiling five independent ways
(\Cref{tab:walls}). (1) An unbounded full-enumeration vote, the solver with no token
limit, answers $457/659 = 0.693$; $25.8\%$ of instances are genuinely ambiguous given the
examples. (2) The hidden map is information-free: $\mathrm{MI}(\text{digit};\text{glyph})
=0.021$ nats versus a shuffle baseline of $0.022$, and a battery of structural predictors
(ascii-offset, pool-linear, first-appearance order, frequency argmax) all score \emph{at or
below} the analytic chance rate of $0.069$ on leave-one-digit-out prediction; the two
``gap'' strata that would require exploiting map structure recover $0$ and $0$ rows. (3) The
base model is structurally lost: native greedy decoding answers $0/50$ and truncates on
$49/50$. (4) The forward derivation is enumeration-locked: a closure that applies only the
steps the base provably executes (column arithmetic, injectivity) covers $1/659$ instances,
and gold maps stall at $0$--$2$ of $\sim\!10$ digits on $449/450$ rows. (5) Per-line
execution fidelity (next paragraph) caps faithful imitation independently. Crucially these
bound \emph{different} quantities: wall 1 bounds the \emph{solver} (any method, $\approx0.69$),
walls 1--2 explain why no map-structure shortcut exists, wall 3 bounds the \emph{base}'s
native behavior, and walls 4--5 bound the only quantity SFT can actually imitate, the
\emph{faithful forward CoT}, near zero. It is that last, binding bound (forward-derivability
$1/659$ together with verdict fidelity) that pins my runs at $\approx0.05$; the solver
ceiling merely says a non-forward, search-based method could in principle reach $\approx0.69$.
The convergence is across \emph{kinds} of bound, not one re-measured number.

\begin{table}[tbp]
\centering
\caption{Five independent bounds on cryptarithm-deduce, each a different measured quantity.
They converge: the task is not learnable as a forward CoT on this base.}
\label{tab:walls}
\small
\begin{tabular}{r l l}
\toprule
\# & Wall (what it measures) & Number \\
\midrule
1 & Solver ceiling (unbounded enumeration vote) & $457/659 = 0.693$; $25.8\%$ ambiguous \\
2 & Information floor (map mutual information) & $0.021$ nats $\approx$ shuffle; gap-strata recover $0$ \\
3 & Base behavior (native greedy) & $0/50$ correct; $49/50$ truncate \\
4 & Forward derivation (provable-step closure) & $1/659$ covered \\
5 & Verdict fidelity (per-line audit) & arithmetic $97$--$100\%$, verdicts $16$--$57\%$ \\
\bottomrule
\end{tabular}
\end{table}

\paragraph{RLVR is alive but does not transfer.} Reinforcement learning against the
verifier behaves exactly as the walls predict. On an \emph{eased} curriculum it works: a
concat-only tier (pure transcription, no arithmetic) reaches mean reward $0.742$ and
$\mathrm{pass@}8 = 15/15$, and a smoke test confirms non-degenerate advantages, the
gradient mechanism is intact. But on real value-operator instances the reward is
$0.000$--$0.004$, and $\mathrm{pass@}16$ on real deduce rows stays near zero across
checkpoints ($0.16, 0.10, 0.05$; $n{=}20$, $95\%$ CIs $\pm 11$--$15$pp and mutually
overlapping, no significant trend, only a persistent floor). A budget-unbounded clean
$\mathrm{pass@}32$ is $0.125$ ($n{=}40$, CI $[0.06,0.26]$). There are essentially no positive
rollouts to learn from, so the held-out accuracy never moves: the gradient is present, but it
has nothing to grip.

\paragraph{Derivation-blocked, not ranking-blocked.} The cleanest single diagnostic is a
generate-and-verify reframe: instead of deriving the map, the model proposes hypotheses and
a learned prior ranks them. Here $\mathrm{hit@}8 = 0.71$ (the gold answer is in the model's
top-8 on $71\%$ of instances) and the worst-case token cost of checking eight hypotheses is
well under budget, so \emph{ranking is not the constraint}. When I instead require the
model to \emph{derive} each hypothesis by forward arithmetic closure, coverage on the
top-8 collapses to $1/659$ and the projected yield ($0.205$) falls below my render gate
($0.45$), so the round was never trained. The model can recognize a correct map but cannot
\emph{construct} one left-to-right, because constructing it requires tracking which
assignments are still consistent, explicit search state, across the whole problem.

\paragraph{Verdict-as-token.} The mechanism becomes concrete in a per-line fidelity audit
of $100$ held-out transcripts ($7{,}566$ line records), scored on three axes: transcription,
arithmetic, and \emph{verdict} (does a line's stated conclusion follow from its own
numbers?). Arithmetic is essentially perfect, the dominant elimination line type
(``enumerate ones-residues, none matches, drop this candidate''; $n=390$) has $100\%$
arithmetic correctness, yet its \emph{verdict} is correct only $34.9\%$ of the time: the
model drops candidates that, by its own numbers on the same line, actually match. Other
elimination types show the same split ($A{=}100\%, V{=}16.5\%$; $A{=}97\%, V{=}51\%$). The
median trace derails at its $552$nd token. The model has learned the \emph{form} of a
verifiable elimination step as a fixed template, decoupled from the logic the step is
supposed to encode. Under teacher forcing this template is always emitted in a context
where it happened to be correct. At free-running inference it fires unconditionally, a
textbook exposure-bias failure \citep{bengio2015scheduled, ranzato2016sequence}, but
localized to the single most load-bearing token in the procedure.

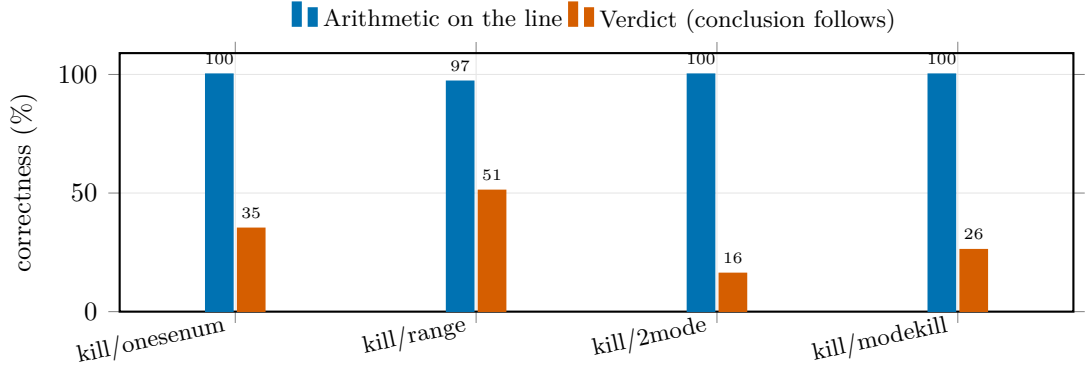
\begin{figure}[tbp]
\centering
\begin{tikzpicture}
\begin{axis}[
  width=0.86\textwidth, height=5.0cm,
  ybar, bar width=10pt, ymin=0, ymax=109,
  ylabel={correctness (\%)},
  symbolic x coords={kill/onesenum, kill/range, kill/2mode, kill/modekill},
  xtick=data, x tick label style={rotate=12,anchor=east,font=\footnotesize},
  legend style={at={(0.5,1.04)},anchor=south,legend columns=2,draw=none,font=\footnotesize},
  nodes near coords, nodes near coords style={font=\tiny,/pgf/number format/fixed,/pgf/number format/precision=0},
  enlarge x limits=0.16,
]
\addplot[fill=c0,draw=c0] coordinates {
  (kill/onesenum,100) (kill/range,97) (kill/2mode,100) (kill/modekill,100)};
\addplot[fill=c1,draw=c1] coordinates {
  (kill/onesenum,35) (kill/range,51) (kill/2mode,16) (kill/modekill,26)};
\legend{Arithmetic on the line, Verdict (conclusion follows)}
\end{axis}
\end{tikzpicture}
\caption{\textbf{Verdict-as-token.} Across the elimination-line types that drive
cryptarithm, the arithmetic on each line is essentially perfect while the line's
\emph{verdict}, ``therefore drop this candidate'', follows from its own numbers only
$16$--$51\%$ of the time. The model imitates the shape of a verifiable step without its
content. Line counts $n=390/102/79/35$. The arithmetic-vs-verdict gap is significant for
every type ($95\%$ CIs $\pm5$--$14$pp).}
\label{fig:fidelity}
\end{figure}

\subsection{The solvability--learnability gap, quantified}
\label{sec:res-gap}

\Cref{tab:master} and \Cref{fig:thesis} make the headline claim precise. The correlation
between solver accuracy and model accuracy is high among the lookup/fit tasks and
\emph{breaks} exactly where the procedure stops being a forward computation and becomes a
search: bit-manipulation (solver $0.99$, model $0.68$) and cryptarithm-deduce (search-solver
$0.71$, model $0.05$). The information-limited ``guess'' tasks are the control: there the
solver itself is low (the answer is not determined by the prompt), and the model tracks it
closely ($0.18$ vs $0.21$; $0.02$ vs $0.59$ where the guess is bottlenecked by the
upstream cryptarithm search). Solver coverage, in short, predicts model accuracy only when
the solver is a forward map. For search procedures it predicts almost nothing, unless the search
is removed from the trace by memorizing its structure (\Cref{sec:memo}).

\section{Anatomy of the Failures}
\label{sec:anatomy}

I characterize the failures at the line and token level across all nine tasks
(\Cref{tab:failuremodes}. Full per-sub-category numbers in \Cref{app:failstats}). Two facts
frame everything. First, \emph{for the shipped adapters, failure is wrong-answer, not
budget}: cryptarithm-deduce finishes within budget and is wrong on $\approx\!96\%$ of
instances ($1\%$ truncated), and bit-manipulation truncates on $0\%$ ($17\%$
finished-but-wrong). The levers that matter are correctness levers, not length levers. The
one exception is diagnostic in itself: the rendered cryptarithm \emph{search} traces
(backtracking narrated step by step) loop and truncate on $64$--$72\%$ of instances, because
the procedure needs a tried-set the model does not maintain. Second, \emph{the errors are
not arithmetic}: a 28-output-bit autopsy and a $7{,}566$-line cryptarithm fidelity audit both
find transcription and arithmetic essentially correct, what fails is the \emph{decision}:
which rule, which verdict.

\paragraph{Bit-manipulation: the failures are rule-selection on hard taps.} Accuracy is a
clean function of tap count (\Cref{tab:bittaps}): one- and two-tap rules are largely solved,
three-tap rules sit at chance-corrected $0.50$. An autopsy of $28$ wrong output bits across
$12$ transcripts found \emph{zero} transcription or arithmetic errors; $21$ were fixable
rule-selection slips (a dropped shift-anchor that shifts a run by one, run accept/reject
steps whose lines did not restate the evidence that should have decided them, and
tie-breaks), and $7$ were genuine three-tap bits for which no $\leq\!2$-input rule exists.
Bit accuracy is therefore bounded by $\approx\!0.82$ in any per-bit grammar and by $\approx\!0.50$
on three-tap rows in any grammar the base can actually execute, which is exactly where STaR
plateaus.

\begin{table}[htbp]
\centering
\caption{Bit-manipulation accuracy by tap count (best adapter, disjoint 500-row eval).
Difficulty is monotone in the number of input taps the rule combines; the three-tap stratum
is the residual ceiling.}
\label{tab:bittaps}
\small
\begin{tabular}{l r l r}
\toprule
Rule complexity & $n$ & Accuracy (95\% CI) & Share \\
\midrule
1 tap  & 42  & 0.881 [.75, .95] & small \\
2 taps & 283 & 0.763 [.71, .81] & $\approx$56\% \\
3 taps & 169 & 0.503 [.43, .58] & $\approx$34\% \\
\midrule
overall & 500 & \textbf{0.678 [.64, .72]} &, \\
\bottomrule
\end{tabular}
\end{table}

\paragraph{Cryptarithm: the verdict is decoupled from the evidence.} The fidelity audit
(\Cref{fig:fidelity}) localizes the failure to elimination \emph{verdicts}. The signature is
concrete and reproducible: the model enumerates candidate residue pairs, \emph{writes one
that matches the target on the same line}, and then concludes ``none matches $\to$ drop.''
The first false line arrives at a median of token $552$. When I instead render the search
explicitly so the model must run it, it either loops forever ($69\%$ of the
generate-and-verify round re-enter an identical attempt, having no record of what was tried)
or, when forced to terminate, teleports to a guess ($31\%$, correct $1/31$). Both are the
absence of carried search state, seen from two sides.

\paragraph{The native-voice attractor (a starkest-case exposure bias).} A distinct and, I
think, broadly important mode surfaced when I authored grammars in a \emph{non-native}
voice (traces opening with \texttt{Bit\dots}/\texttt{Group\dots} rather than the base's
habitual \texttt{I\dots}). Teacher-forced fidelity on these grammars was
$0.92$--$0.95$, the model demonstrably \emph{learned} them, yet at greedy decoding it
entered them \emph{$0\%$ of the time}: at the very first generated token it emits its native
opener instead, a $15$--$16$ nat divergence, and never reaches the learned procedure. The
grammar is a fully-learned but unreachable island. Cryptarithm, whose traces happen to share
the native opener, does enter the grammar and then \emph{drifts} mid-trace, which is how it
manages to score \emph{below} a constant-output baseline (destructive interference with the
base's own attempt). The operational lesson is sharp: author chain-of-thought in the base's
native voice, or it will never be decoded at inference no matter how well it is fit. This is
exposure bias \citep{bengio2015scheduled, ranzato2016sequence} at its most extreme, the
training and free-running distributions diverge at token $0$.

\begin{table}[htbp]
\centering
\caption{The nine observed failure modes collapse to four root causes; the first, a verdict
decoupled from its own evidence, dominates, and is the unfaithful-CoT signature
\citep{turpin2023unfaithful, lanham2023measuring} in the training regime.}
\label{tab:failuremodes}
\small
\begin{tabularx}{\linewidth}{@{}p{2.9cm} X X@{}}
\toprule
Root cause & Variants (dominant category) & Measured signature \\
\midrule
1. Verdict decoupled from evidence (``verdict-as-token'') & teleport / unwitnessed claim. Verdict-as-constant; fabrication-on-exhaustion; constant-concat rubber-stamp (crypt; authored bit) & verdict fidelity 16--51\% vs arithmetic 97--100\%; SFT rounds capped 0.03--0.07 \\
2. No carried search state & infinite loop / no tried-set. Budget-truncating search renders (crypt) & re-enters identical attempts; loops to 7680-tok cap on 64--72\% \\
3. Distribution mismatch (exposure bias) & never-enters / anchoring. Mid-trace drift (non-native grammars; crypt) & teacher-forced fidelity 0.95 yet greedy adoption 0.0; diverges at/after token 0 \\
4. Output corruption & symbol-table hallucination (crypt encode) & correct derivation, glyphs emitted out of order (T fidelity 51\%) \\
\bottomrule
\end{tabularx}
\end{table}

\section{Analysis: Why Some Procedures Resist Distillation}
\label{sec:analysis}

My results compose into one account, in three threads. \emph{Thread 1} is a controlled
refutation: distilling the search fails on every backbone, not just the hybrid one (every
backbone caps). \emph{Thread 2} identifies the failure mechanism: SFT installs the surface form of
a verifiable step, a \emph{verdict decoupled from its evidence}. \emph{Thread 3} gives the root
reason: the search admits no faithful forward chain-of-thought to imitate. The chain is: no
faithful forward trace $\Rightarrow$ SFT copies only the shape $\Rightarrow$ verdict-as-token
$\Rightarrow$ collapse at free-running inference, on every backbone. The dual escape, memorizing
the search's finite structure rather than distilling the search, is taken up in \Cref{sec:memo}.

\paragraph{1. The bottleneck is the task, not the architecture.}
The base model has every primitive cryptarithm needs, it does column arithmetic correctly
$97$--$100\%$ of the time, recognizes a correct map $71\%$ of the time when one is proposed,
and handles the purely transcriptional concat family. What it cannot do is run a search:
maintain the set of tried assignments, propagate constraints, and decide a branch is dead.
I first suspected the hybrid Mamba-2/MoE backbone: state-space layers compress history into a
fixed-size state \citep{dao2024mamba2} that sits in $\mathrm{TC}^0$ and cannot perform genuine
sequential state-tracking \citep{merrill2024illusion}, a natural reason a growing search
frontier cannot be carried. A controlled architecture sweep \emph{refutes} that explanation for
this task (\Cref{tab:archctrl}): trained on the identical cryptarithm corpus, two dense
Transformers (Llama-3.2-3B, Qwen3.5-4B) and an MoE Transformer (gpt-oss-20b) hit the \emph{same}
$\leq\!0.04$ floor, each fits the corpus (train NLL $\to$ floor) yet reproduces almost none of
it at inference. The search-distillation failure is therefore \emph{architecture-general}, and
the binding constraint is task-intrinsic, no faithful forward CoT to imitate (thread 3), not the
SSM backbone. (State compression may still contribute to the bit-manipulation three-tap tail,
where a rank-$\leq\!32$ adapter is too small a perturbation to install a search memory.) The same
holds \emph{without} fine-tuning: sampled in-context under the competition's $7680$-token budget,
DeepSeek-V3.1 ($671$B) scores $0.05$ (finishes, wrong) and Nemotron-Super-$120$B scores $0.00$
(truncates on $100\%$), so it spans $3$B$\to$$671$B and SFT$\to$in-context alike
(\Cref{tab:archctrl}).
The bit autopsy is the same story in miniature: of $28$ wrong output bits, $0$ were transcription or
arithmetic errors. All traced to run-extension \emph{verdicts} (accept/reject/tie-break)
whose lines did not restate the evidence that should have decided them.

\begin{table}[htbp]
\centering
\caption{\textbf{The search does not distill, on any backbone.} Cryptarithm-deduce accuracy, all under
the competition's $7680$-token budget. \emph{Top}: the identical corpus fine-tuned (rank-32
LoRA, 4 epochs) on four backbones, $3$B--$30$B. \emph{Bottom}: two frontier models sampled
\emph{in-context} (no fine-tuning. The prompt already carries the worked examples). Spanning
$3$B$\to$$671$B and fine-tuned$\to$in-context, none reproduces the search. At frontier scale the
failure splits: V3.1 finishes but is wrong. The 120B truncates on $100\%$ within budget. The
exact 30B/A3B Transformer match (Qwen3-30B-A3B) was unschedulable on my compute.}
\label{tab:archctrl}
\small
\begin{tabular}{l l l c l}
\toprule
Model & Params (tot/act) & Architecture & deduce & note \\
\midrule
\multicolumn{5}{@{}l}{\itshape Fine-tuned (rank-32 LoRA, identical corpus; $n{=}100$)} \\
Nemotron-3-Nano & 30B / 3.5B & Hybrid Mamba-2+MoE & 0.03--0.05 & \\
Llama-3.2-3B    & 3B / 3B    & Dense Transformer  & 0.010 & \\
Qwen3.5-4B      & 4B / 4B    & Dense Transformer  & 0.040 & \\
gpt-oss-20b     & 21B / 3.6B & MoE Transformer    & 0.010 & \\
\addlinespace
\multicolumn{5}{@{}l}{\itshape In-context, no fine-tuning (raw base; $n{=}20$)} \\
DeepSeek-V3.1   & 671B / 37B & MoE Transformer    & 0.050 & finishes, wrong \\
Nemotron-Super  & 120B / 12B & Hybrid Mamba+MoE   & 0.000 & 100\% truncate \\
\bottomrule
\end{tabular}
\end{table}

\paragraph{2. Verdict-as-token is the failure's signature under teacher forcing.} A
verifiable solver's most important lines are its verdicts, ``this candidate is impossible,''
``this run extends.'' When a CoT is written so that a verdict is a fixed phrase rather than a
function of locally-written evidence, SFT learns the phrase, not the function, because under
teacher forcing the phrase is always seen in a correct context. The defect is invisible in
training loss, across fourteen cryptarithm training rounds the train NLL converged to the
floor (e.g.\ crypt-r1 reaches NLL $0.146$ by step $35$ and $\approx\!0.001$ by the end; the
Kaggle merge runs log a final loss $\approx\!0.003$) while held-out deduce accuracy never
left $[0.01,0.07]$ (\Cref{tab:ledger}), and invisible under teacher-forced evaluation. It appears only at free-running
greedy decode, as the verdict firing in the wrong context. This is why every \emph{authored}
trace failed and only \emph{self-harvested} traces (STaR) succeeded for bit-manipulation:
a self-trace's verdict is, by construction, one the model produced \emph{and the verifier
confirmed}, so the evidence$\to$verdict link is real rather than narrated.

\paragraph{3. The root cause: no faithful forward CoT to imitate.} Cryptarithm
is the extreme case: the hidden map is information-free (wall 2), so there is no
left-to-right derivation that a human or solver could write down that arrives at the map by
forward reasoning, the only honest route is backtracking search, which forward CoT cannot
represent without carried state. The forward-closure coverage of $1/659$ is the formal
statement: the set of instances solvable by the steps the base can actually execute,
in order, is empty for all practical purposes. When no faithful forward CoT exists, SFT can
only learn an unfaithful one, and the unfaithful one is exactly the verdict-as-token
template that collapses at inference.

\paragraph{A predictive rule of thumb.} Together these suggest a cheap pre-test for whether a
solvable task will be CoT-learnable on a small model, applied \emph{before} any training:
(a) does a purely forward closure (no backtracking, no global constraint propagation) cover a
meaningful fraction of instances? (b) does the task's hidden structure carry non-trivial
mutual information with observable tokens? (c) does the base model, prompted natively,
attempt the right \emph{kind} of move rather than truncating? Cryptarithm fails all three;
bit-manipulation passes (a) and (c) and is learnable via the model's own search; the easy
four pass trivially. I derived this \emph{post-hoc} from the nine tasks here, but test its core claim, that
forward-derivability is the causal lever, prospectively next (\Cref{tab:interv}); the broader
three-part heuristic remains a hypothesis I offer for testing.

\paragraph{A prospective test: the forward-derivability intervention.} To check the core claim
causally rather than by correlation, I took the \emph{same} cryptarithm instances and revealed
a fraction of the digit$\leftrightarrow$symbol key in the prompt, turning that fraction of the
derivation from search into forward lookup, then fine-tuned Nemotron on each (\Cref{tab:interv};
$n{=}42$). With no key (the original task) accuracy is $0.03$; revealing \emph{half} the key
barely moves it ($0.048$). Revealing the \emph{full} key lifts it to $\mathbf{0.571}$, a
$\sim\!15\times$ jump on an otherwise identical task (disjoint $95\%$ CIs), with the residual gap
to $1.0$ explained by ordinary op-execution errors (cf.\ equation\_numeric at $0.83$). Two things
follow. First, forward-derivability is \emph{causal}: it is the lever, holding the surface task
fixed. Second, the effect is sharply \emph{nonlinear}, half-forward does not give half-accuracy,
because any residual search re-introduces the verdict-as-token failure; a faithful forward CoT
must cover the \emph{whole} derivation, not most of it.

\begin{table}[htbp]
\centering
\caption{\textbf{Forward-derivability is causal.} The \emph{same} cryptarithm instances, with a
fraction of the cipher key revealed in the prompt (turning that fraction of the derivation
forward), fine-tuned on Nemotron ($n{=}42$, Wilson 95\% CI). Making the task fully forward lifts
accuracy $\sim\!15\times$. Half-forward barely helps, any residual search re-triggers the
collapse.}
\label{tab:interv}
\small
\begin{tabular}{l c c}
\toprule
Key revealed & Forward-derivable & crypt-deduce \\
\midrule
none (original task) & $\approx$0 & 0.03 \\
half & $\approx$50\% & 0.048 [.01, .16] \\
full & 100\% & \textbf{0.571} [.42, .71] \\
\bottomrule
\end{tabular}
\end{table}

\section{Competition Dynamics and External Corroboration}
\label{sec:dynamics}

\subsection{Leaderboard structure and external corroboration}

The benchmark is drawn from a public competition \citep{nemotronchallenge2026}, whose final
$4{,}355$-team leaderboard is itself evidence for my thesis. The score distribution has a
sharp plateau exactly where I place the frontier: $2{,}236$ teams reached $\geq\!0.85$, only
$66$ cleared $0.87$, $7$ cleared $0.88$, and just two teams reached $\geq\!0.90$, with a
single leader at $0.92$. A score of $0.85$ is the \emph{median}. This is the shape my
analysis predicts: the easy four plus a learnable share of bit-manipulation get a competent
team to $\approx\!0.85$, and the last few points require cracking cryptarithm, which almost
no one did.

\paragraph{Independent corroboration of the ceiling.} The competition's Open Progress Prize
(awarded for the best reproducible \emph{methodology}, separately from the raw leaderboard)
went to a fully open-source solution \citep{huikang2026nemotron, huikang_kaggle2026} that,
independently of me, took the \emph{same} approach, reverse-engineer each generator, then SFT a rank-$32$ LoRA on
solver-grounded synthetic chain-of-thought (LoRA rank $32$, adapt MLP/attention/unembed, LR
$2\times10^{-4}$, single epoch, $8192$ context, category-stratified batching). It also scored
$\mathbf{0.85}$. Two independent teams converging on the same method \emph{and} the same
ceiling is strong evidence that $0.85$ is the ceiling of the \emph{search-distillation approach}
both used, not of either team's execution; the 1st-place solution's memorization--computation
reformulation breaks past it to $0.92$ (\Cref{sec:memo}). My training recipe and theirs match
closely (\Cref{app:training}), which is no coincidence: ours was tuned toward the same proven regime.

\paragraph{The residual is generator hardness, not modeling laziness.} My own frontier
baseline agrees: DeepSeek-V3.1 ($671$B) and Nemotron-Super-$120$B, prompted in-context on
cryptarithm, score $0.05$ and $0.00$ (\Cref{tab:archctrl}). And a third-party effort reports
several hundred of the hardest puzzles \emph{unsolved even by frontier teacher models} given the
recovered rule at high reasoning effort. The ceiling is in distilling the search, not in a failure to try.

\subsection{Threats to validity}
Three are worth recording. (1) \emph{The metric has a
binary-string trap}: binary answers are compared as exact strings, so a correct value at the
wrong width (\texttt{"11011"} vs \texttt{"00011011"}) is scored wrong; this is a hazard to
avoid (emit the exact width), not a lever. (2) \emph{Public/private overfit}: competitors
discussed choosing a cryptarithm default operation (reverse- vs.\ forward-concatenation) that
raised the \emph{public} score while \emph{lowering} the private one, a leaderboard-specific
artifact, and a caution for anyone reading single-split numbers. (3) \emph{Memorization under
train$\equiv$test}: with train and test from one generator, multi-epoch training can drive
the training metric to ceiling by memorizing rather than learning the procedure, which is
precisely why I train on synthetic instances with \emph{different} values and report
held-out (``unseen'') accuracy rather than the contaminated headline (\Cref{app:failstats}).
A concrete leaderboard-gaming technique I can \emph{document} rather than merely report is
\textbf{adapter-noise laundering}: a notebook in my own repository loads a \emph{third
party's} shared submission adapter and adds Gaussian noise at $20\%$ of each tensor's standard
deviation to all $12{,}010$ weight tensors, then repackages it as a fresh submission. The
perturbation is small enough to preserve most of the donor's accuracy while yielding a
nominally distinct artifact. I did not measure its leaderboard effect (my copy records no
score), so I report the \emph{procedure} as a reproducible integrity concern, not a
result, a shared high-scoring adapter can be re-skinned into many ostensibly independent
submissions.

\paragraph{My own public/private near-miss (verified).} The selection risk was not
hypothetical for me. My full submission record (Kaggle API; \Cref{tab:pubpriv}) shows my
best \emph{private} score was $\mathbf{0.860}$, a native-HuggingFace run (EXP-138, forking
Tong's published recipe \citep{huikang_kaggle2026}) trained on the $0.86$-adapter's
verifier-passed traces plus $1{,}600$ synthetic bit puzzles, yet that submission scored only
$0.844$ \emph{public}, while my public-best submissions sat at $0.852$--$0.856$. The trap is
sharpest in the reference adapter I benchmarked: $0.856$ public but $0.832$ private (it
overfit the public split). Standard public-LB selection would have discarded my
$0.860$-private model and kept a weaker one. The public score, computed on a few-hundred-row
split, is a lossy selector. A genuinely stronger model can go unselected, which is how a
medal-class private result was left on the table.

\begin{table}[htbp]
\centering
\caption{Public vs.\ private leaderboard for my key submissions (Kaggle API record). The
ordering inverts: my best \emph{private} score ($0.860$, EXP-138) is mid-pack on
\emph{public}, while the best \emph{public} score (the reference adapter, $0.856$) is among the
worst on private. Selecting on the public column picks the wrong submission.}
\label{tab:pubpriv}
\small
\begin{tabular}{l r r}
\toprule
Submission & public & private \\
\midrule
EXP-138 (native-HF, 06-05)        & 0.844 & \textbf{0.860} \\
v18 ($+$ bit STaR)                & 0.848 & 0.856 \\
(06-06)                           & 0.840 & 0.852 \\
v17 (banked)                      & 0.852 & 0.852 \\
v15                               & 0.828 & 0.840 \\
reference adapter ($\approx$0.86) & \textbf{0.856} & 0.832 \\
v6\_7                             & 0.768 & 0.768 \\
\bottomrule
\end{tabular}
\end{table}

\section{Memorization, not search}
\label{sec:memo}

The search-distillation ceiling documented above is a ceiling on \emph{distilling the search},
not on the task. The competition's 1st-place solution (team NullSira; Private LB $0.920$) reaches high
cryptarithm and bit-manipulation accuracy by an approach its authors describe as deciding
``what the model should memorize \dots\ and what it should compute inside the trace''
\citep{goodmeatday2026first}, and it is the cleanest illustration of the present thesis: it
succeeds precisely by \emph{not} teaching the model to search.

The naive search over a cryptarithm prompt is intractable to narrate, up to $10!$ digit
assignments times $24^3$ operator rules ($\approx\!5\times10^{10}$ candidates), which cannot be
enumerated within the $7680$-token budget. The winning solution removes that search from the
trace. For a single equation it precomputes a \emph{signature}, the pattern of repeated symbols
normalized by first occurrence (e.g.\ \texttt{ABCCCDD}), and enumerates all two-digit
$\times$ two-digit operands under the $22$ non-join rules into a catalog of $4{,}205$ signatures,
each mapping to a short list of candidate (rule, digit-assignment) rows with counts. The model
\emph{memorizes this catalog} through heavy SFT; at inference it recalls the candidate list for
each equation rather than deriving it. The trace then does only the cheap residual: pick the
equation with the fewest candidates as an anchor, and run a bounded DFS that checks the recalled
candidates against the remaining equations (same symbol $\to$ same digit, same operator $\to$ same
rule), pruning by mode and group constraints. The combinatorially hard step, candidate generation
for the first equation, is memorized; only a short, bounded verification remains as
chain-of-thought.

Bit-manipulation is handled the same way: $5{,}238$ valid $8$-rule sequences (every coherent
$8$-bit transformation, normalized to per-output-bit rules) are precomputed and memorized, the
greedy per-bit search is projected onto the nearest catalog sequence under Hamming distance, and
candidates are verified against the examples (bit columns HEX-compressed to fit the budget). In
both cases the pattern is the dual of my negative result. Distilling the search fails because no
faithful forward trace exists (verdict-as-token); the escape is to compute the search
\emph{offline} into a finite catalog, store it in the weights as memorized lookup, and leave only
recall plus bounded verification in the trace. What transfers is therefore not search but
\emph{memorization and verification}. This sharpens the boundary beyond ``solvable vs.\
learnable'': a verifiable search is learnable as a chain-of-thought exactly when its candidate
structure compresses to a catalog small enough to memorize and a residual cheap enough to verify
in-budget. (The winners' solver-coverage figures include train-set leakage by their own account,
but their $91.57\%$ held-out validation, on the same train$\equiv$test generators I use, confirms
generalization within the distribution rather than row memorization.)

\section{Limitations}
\label{sec:limitations}

My study is deep on one benchmark rather than broad. I \emph{did} run a controlled
architecture sweep (\Cref{tab:archctrl}): training the identical cryptarithm corpus on dense
Transformers (Llama-3.2-3B, Qwen3.5-4B) and an MoE Transformer (gpt-oss-20b) yields the same
$\leq\!0.04$ floor as the hybrid, so the search-distillation ceiling is architecture-general
rather than Mamba-specific.
On scale, frontier models up to $671$B sampled in-context under the competition budget also cap
($0.05$ / $0.00$, \Cref{tab:archctrl}). The one cell I could not run is a frontier model
\emph{fine-tuned} at length (and the exact 30B/A3B Transformer match, unschedulable on my
compute). I also report single training seeds per cell. The binomial CIs bound sampling noise
but not seed variance. I evaluate under greedy decoding because the grader does; sampling-based test-time
search would change the achievable numbers (my $\mathrm{hit@}8$ and $\mathrm{pass@}32$
results quantify by how much) but not the single-trace learnability question I study. Some
internal figures I deliberately did not promote into the headline claims because they
appear in my run ledger but not in a re-runnable report (e.g.\ a $\mathrm{pass@}32$ value
attached to a specific value-rule count); I cite only the report-backed numbers and flag
this in my released notes. Finally, I did not win the competition: my best \emph{public}
score was $0.852$ and my best \emph{private} score $0.860$, medal-class, but on a submission
public-LB selection would not have chosen (\Cref{sec:dynamics}); the leader reached $0.92$.
I regard that as orthogonal to the contribution, the value here is the instrumented
account of the frontier, not the rank. And the cryptarithm ceiling \emph{does} yield to an idea I
did not pursue: the memorization--computation split of the 1st-place solution (\Cref{sec:memo}).
My contribution is the mechanistic account of why distilling the search fails, which is precisely
why that split is necessary.

\section{Conclusion}
\label{sec:conclusion}

I set out to convert perfect symbolic solvers into a small model's chain-of-thought, on a
benchmark engineered so that held-out training accuracy is the test score. Most tasks
transferred. One did not, and its resistance was not a matter of effort: eleven CoT designs,
RLVR, and STaR all left cryptarithm at $\approx\!0.05$, and five independent measurements
agreed on why. The unifying lesson is that a verifiable \emph{search} is not a learnable
chain-of-thought. What a small autoregressive model can imitate is a forward computation;
what a search procedure requires is carried state, a record of what has been tried and a
decision about when to stop, and when a task's only honest solution is search, SFT can
learn only the \emph{shape} of the solver's verdicts, which then misfire at inference. There are
two ways across this gap, and both avoid distilling the search itself: teach the model a search it
can actually run (bit-manipulation, via STaR on its own successful traces), or remove the search
from the trace entirely, precomputing its finite structure into a catalog the model memorizes and
verifies against (\Cref{sec:memo}; the 1st-place solution). The right way to read solver coverage,
then, is as a necessary but badly insufficient condition: it tells you the answer exists, not that
your model can reason its way to it, and least of all that the solver's \emph{search} will become
its chain-of-thought.

\section*{Ethics and Broader Impact}
This work is diagnostic and defensive: it introduces no new model capability and no attack on
a deployed system. It does surface competition-integrity concerns the community should weigh:
a public/private leaderboard split makes the public score a lossy selector (\Cref{sec:dynamics}),
and a shared high-scoring adapter can be trivially perturbed (``adapter-noise laundering'')
into many ostensibly independent submissions. I report these to inform benchmark and
competition design (private-split selection, submission provenance), not to enable
gaming, I did not submit the noised adapter as my own work. Finally, the
deterministic-generator benchmark is a controlled proxy for procedural reasoning; my
findings should not be read as a claim about a model's general capability.

\section*{Reproducibility and Artifacts}
Code, paper source, the architecture-control scripts, and per-row eval data are released at
\url{https://github.com/harshpatel1692/search-not-learnable} (interactive walkthrough at
\url{https://nemotron.harshpatel.live}). I release, per category: the reverse-engineered solvers
and the synthetic-CoT renderers; the held-out evaluation harness; the per-row eval CSVs behind
the figures; and the per-line fidelity audit ($7{,}566$ records). The benchmark data itself
belongs to the competition and is available from Kaggle, not redistributed here. Numbers in this
paper are drawn from a dated experiment log. Where a figure exists only in my run ledger and not in a
re-runnable report, I say so rather than promote it.

\appendix
\section{Solver Pseudocode}
\label{app:solvers}

Each solver is the data-generation oracle for its category: it recovers the hidden rule
from the worked examples and applies it to the query. ``Coverage'' is the fraction of the
$9{,}500$ training rows the solver answers correctly. I give the four lookup/fit solvers
compactly, then the two search solvers (equation and cryptarithm) in full; the
bit-manipulation solver is in \Cref{app:bit}.

\paragraph{The four forward solvers (numeral, unit, gravity, cipher).}
\emph{numeral} (100\%) is a fixed greedy subtractive encoder over the 13 Roman value-symbol
pairs, it ignores the examples entirely. \emph{unit\_conversion} (100\%) fits the linear
map $y=ax+b$ from the two most-separated example points (for numerical stability under the
grader's $10^{-2}$ tolerance) and applies it. \emph{gravity} (100\%) fits the single
constant $A=d/t^2$ from one example of $d=\tfrac12 g t^2$ and applies $A\,t_*^2$. Only
\emph{cipher} requires search, over a \emph{closed} vocabulary; I give it in
\Cref{alg:cipher}.

\begin{algorithm}[htbp]
\caption{\textsc{Cipher}, monoalphabetic substitution over a closed 77-word vocabulary (100\%)}
\label{alg:cipher}
\begin{algorithmic}[1]
\Require example pairs $\{(c_i,p_i)\}$ (cipher word $\to$ plain word); query cipher words $Q$;
  global plaintext vocabulary $V$, $|V|=77$
\State $m \gets \{\}$;\ $\mathit{inv}\gets\{\}$ \Comment{cipher$\to$plain letter map and its inverse}
\For{each pair $(c_i,p_i)$ with $|c_i|=|p_i|$} \Comment{seed from equal-length alignments}
  \For{each position $j$} $m[c_i[j]]\gets p_i[j]$;\ $\mathit{inv}[p_i[j]]\gets c_i[j]$ \EndFor
\EndFor
\Repeat \Comment{constraint propagation to fixpoint}
  \For{each query word $w\in Q$ not fully decided}
    \State $\mathit{cand}\gets\{u\in V : |u|=|w|,\ u \text{ consistent with } m,\mathit{inv}\text{, and intra-word injective}\}$
    \If{$|\{\,u : u\in\mathit{cand}\,\}|=1$} lock that word's letters into $m,\mathit{inv}$ \EndIf
  \EndFor
\Until{no change}
\State \Return $\textstyle\bigsqcup_{w\in Q}$ decode$(w,m)$ \Comment{unresolved letters render as \texttt{?}}
\end{algorithmic}
\end{algorithm}

\paragraph{Equation\_numeric (deduce 95.1\%, guess 20.6\%).} Each example is
$AB\,g\,CD=\textsc{rhs}$ with \emph{visible} digits. The row has one global string mode
(identity or reverse) and each operator glyph $g$ carries one operation from a 13-op library
(\Cref{alg:eq}). The policy locks the query operator to the first op in a
frequency-ordered list consistent with \emph{all} of that glyph's examples, then accepts the
mode only if every other glyph is also explainable under it. For an unseen query glyph
(``guess'') the operation is undeducible, so the policy pins the mode from the known glyphs
and bets the marginal-prior operation (\texttt{sub}), which is exactly the $20.6\%$ marginal
ceiling.

\begin{algorithm}[htbp]
\caption{\textsc{Equation\_numeric}, \texttt{policy\_seq} (deduce 95.1\%, guess 20.6\%)}
\label{alg:eq}
\begin{algorithmic}[1]
\Require examples grouped by glyph $g$; query $(qa,qg,qb)$;
  op library $\mathcal{O}$ (13 ops). Frequency order $\textsc{seq}$; default sign-render table
\For{$\mathit{mode}\in(\textsc{rev},\textsc{id})$} \Comment{reverse first: generator prior $476{:}239$}
  \State $\mathit{qop}\gets$ first $o\in\textsc{seq}$ consistent with \emph{all} examples of $qg$ under $\mathit{mode}$
  \If{$qg$ unseen} $\mathit{qop}\gets\texttt{sub}$ \EndIf \Comment{guess: marginal-prior bet}
  \If{every other glyph $g'$ also has some consistent op under $\mathit{mode}$} \Comment{global-mode check}
    \State \Return \textsc{render}$(qa\ \mathit{qop}\ qb,\ \mathit{mode},\ \textsc{signpos}(\mathit{mode},qop))$
  \EndIf
\EndFor
\end{algorithmic}
\end{algorithm}

\paragraph{Cryptarithm (deduce: policy 67.7\%, posterior vote 70.9\%).} This is the search
solver, and the contrast between it and what a forward CoT can imitate is the heart of the
paper. The hidden per-row digit$\to$symbol cipher is information-free, so no column reveals a
digit. The map is pinned only by a backtracking search that requires \emph{all} example
equations to hold simultaneously, under the right per-glyph operation (from a $\sim$30-op
vocabulary), reading direction, and sign convention (\Cref{alg:crypt}). Two sound prunes
(units-digit congruence and full-operand forcing with injectivity) and a most-constrained
variable order keep the $10!$ assignment space tractable; the solver enumerates up to
$3000$ consistent assignments and scores each under a two-regime (canonical $+$ scrambled)
mixture prior fit by EM, then returns the answer with maximum posterior mass. Crucially, the
\emph{forward-only} variant of this procedure (constraint propagation with no backtracking,
\Cref{alg:cryptfwd}) is what a left-to-right CoT could honestly reproduce, and it covers
$\approx\!1/659$ of instances, which is why every supervised attempt to teach the search
collapses (\Cref{sec:res-crypt}).

\begin{algorithm}[htbp]
\caption{\textsc{Cryptarithm}, search solver \texttt{cryptarithm2} (deduce vote 70.9\%)}
\label{alg:crypt}
\begin{algorithmic}[1]
\Require example equations $\{(L_i,R_i)\}$, each LHS $=s_0 s_1\,\textsc{op}\,s_2 s_3$. Query LHS; op vocab $\mathcal{O}$ ($\sim$30); EM priors
\State $\mathit{mass}\gets\{\}$
\For{$(\mathit{rev},\mathit{radix})\in\{(\bot,10),(\top,10),(\bot,11),(\top,11)\}$} \Comment{radix 11 only as fallback}
  \State normalize rows for $\mathit{rev}$ (reverse magnitudes/result. Detect sign glyph); reject if a sign glyph sits inside a magnitude
  \State per glyph, restrict candidate ops by sign pattern and operand magnitude bounds. If all rows match a structural \texttt{concat}, fix \texttt{concat}
  \State $\mathcal{S}\gets$ DFS over digit$\to$symbol assignments (most-shared symbol first), pruned by units-digit congruence and full-operand forcing $+$ injectivity, up to $3000$ solutions
  \For{each solution $\sigma\in\mathcal{S}$}
    \State $\mathit{score}\gets$ logsumexp of canonical-regime and scrambled-regime log-priors of $\sigma$'s op assignment
    \State render the query answer under $\sigma$; $\mathit{mass}[\text{answer}]\mathrel{+}=e^{\mathit{score}}$
  \EndFor
\EndFor
\State \Return $\arg\max_{a}\mathit{mass}[a]$ \Comment{posterior-mass vote. Ties by max score then lexicographic}
\end{algorithmic}
\end{algorithm}

\begin{algorithm}[htbp]
\caption{\textsc{Cryptarithm-forward}, the imitable forward closure ($\approx\!1/659$ coverage)}
\label{alg:cryptfwd}
\begin{algorithmic}[1]
\Require committed $(\mathit{op},\mathit{rev})$ per glyph; example equations
\State initialize every symbol's digit domain to $\{0,\dots,9\}$ \Comment{no teleporting}
\Repeat
  \State \textbf{units mod 10}: narrow domains via $(x\circ y)\bmod 10$ on placed units symbols
  \State \textbf{magnitude bracket}: narrow via result length / value range
  \State \textbf{all-different cascade}: remove singletons from other domains
  \State \textbf{per-equation GAC}: keep only operand/result digits that appear in some consistent $(a,b)$
\Until{no change \textbf{or} stalled}
\If{every query operand symbol has a singleton domain} \Return forward-derived answer
\Else{} \Return \textsc{stalled} \Comment{the common case: $\approx\!658/659$}
\EndIf
\end{algorithmic}
\end{algorithm}

\section{Bit-Manipulation Solver, Basis, and Renderer}
\label{app:bit}

\paragraph{Global single-rule search (98.88\%).} The production solver
(\Cref{alg:bitglobal}) treats the rule as one boolean function over up to three
\emph{taps} and searches a composed expression grammar rather than the named basis directly.
A \emph{tap} is a transformed copy of the 8-bit input, $\textsc{rot}/\textsc{shl}/\textsc{shr}$
by $k\in\{0,\dots,7\}$, shifts zero-filling at the boundary, giving $22$ taps; \textsc{not}
is folded into the grammar. The grammar enumerates boolean expressions over three projection
variables by op-composition to depth $3$ (ten two-input gates plus unary \textsc{not}),
deduped by 8-bit truth table. For each expression the solver tries every assignment of
distinct taps to its variables (a single var: $22$; two: $22\!\cdot\!21$; three:
$\mathrm{perm}(22,3)=9240$), fail-fast on output bit $0$ before checking bits $1$--$7$, and
accepts the first assignment that reproduces \emph{all} eight bits of \emph{every} example;
it then applies that rule to the query. On the cache of all $1602$ rows the solver finds a
consistent rule for every row but only $1584$ ($98.88\%$) generalize to the query, the
$18$-row gap is genuine over-fitting (a rule consistent on the shown examples that predicts
the wrong query output). The same single-rule space, re-described in human terms, is the
basis of \Cref{tab:basis}; ``coverage by the basis'' ($\approx\!98\%$ \emph{fit on examples})
and ``$98.88\%$'' (\emph{correct on the held-out query}) are different measurements and
should not be conflated.

\begin{algorithm}[htbp]
\caption{\textsc{Bit-manipulation}, global single-rule search (98.88\%, 1584/1602)}
\label{alg:bitglobal}
\begin{algorithmic}[1]
\Require example pairs $\{(\mathit{in}_i,\mathit{out}_i)\}$ of 8-bit strings; query input $x_*$
\State $\mathcal{T}\gets$ the $22$ taps $\{(\textsc{rot},0)\}\cup\{(\textsc{op},k):\textsc{op}\in\{\textsc{rot},\textsc{shl},\textsc{shr}\},k=1..7\}$
\State $\mathcal{G}\gets$ boolean expressions over vars $\{A,B,C\}$, composed to depth $3$ (10 binary gates $+$ \textsc{not}), deduped by truth table
\For{each expression $e\in\mathcal{G}$ in increasing depth}
  \For{each assignment of distinct taps in $\mathcal{T}$ to the variables \emph{used} by $e$}
    \If{$e$ reproduces output bit $0$ of all examples} \Comment{fail-fast}
      \If{$e$ reproduces \emph{all} 8 bits of all examples}
        \State \Return apply $e$ (with this tap assignment) to $x_*$
      \EndIf
    \EndIf
  \EndFor
\EndFor
\State \Return \textsc{none} \Comment{$\approx\!1\%$: under-constrained / $>$3-tap tail; 5\,s time limit}
\end{algorithmic}
\end{algorithm}

\paragraph{Tap-count distribution (from the solve cache).} The $1602$ recovered rules split
$0$-tap $5$, $1$-tap $149$, $2$-tap $899$ ($56\%$), $3$-tap $549$ ($34\%$). The two- and
three-tap shares are exactly the strata in the model's accuracy breakdown
(\Cref{tab:bittaps}). The three-tap stratum is both the largest hard slice and the residual
ceiling. The $18$ over-fit failures (a rule consistent on the examples but wrong on the
query) span $4$ one-tap, $7$ two-tap, and $7$ three-tap rows.

\paragraph{Verifiable rule inventory.} \Cref{tab:bitcache} is the op-composition taxonomy of
all $1602$ recovered rules, read directly off the solve cache, the machine-checkable
complement to the human-named basis of \Cref{tab:basis}. The two views agree on the shape
(\textsc{xor} dominant, a large \textsc{xnor} stratum corresponding to the basis's
\textsc{or\_xnor}/\textsc{choice} families, identity/copy rules, and AND/OR combinations);
the named basis is the interpretable face, this table is the reproducible ground truth.

\begin{table}[htbp]
\centering
\caption{The op-composition rule inventory over all $1602$ training rows, computed directly
from the solve cache (root operator of the recovered expression; multi-input rules compose
these). Taps are near-uniform over rotate/shift by $k\in\{1,\dots,7\}$.}
\label{tab:bitcache}
\small
\begin{tabular}{l r | l r}
\toprule
Root operator & Rows & Tap count & Rows \\
\midrule
\textsc{xor}          & 444 & $0$ taps & 5 \\
\textsc{or}           & 350 & $1$ tap  & 149 \\
\textsc{xnor}         & 288 & $2$ taps & 899 \\
\textsc{and}          & 187 & $3$ taps & 549 \\
identity / copy       & 149 & & \\
\textsc{not\_a\_and\_b} & 96 & tap type & \\
\textsc{not\_a\_or\_b}  & 79 & \textsc{shl} & 1253 \\
constant              & 5   & \textsc{shr} & 1197 \\
\textsc{nor}          & 4   & \textsc{rot} & 1144 \\
\bottomrule
\end{tabular}
\end{table}

\paragraph{Per-bit stride renderer (84.6\%, used only to author CoT).} To \emph{narrate} a
search the model could imitate, a second solver expresses each output bit $j$ as a constant,
a (possibly negated) copy of one input bit, or a two-input gate of two input bits, with the
referenced input positions advancing by a fixed stride as $j$ increases (because taps are
rotations/shifts). It greedily set-covers the eight output bits by maximal cyclic runs,
preferring longer runs and simpler descriptors. Output bits needing three inputs are left
uncovered and default to $1$, the source of its $84.6\%$ ceiling. This renderer is a CoT
authoring tool, not the production solver. As \Cref{sec:res-easy} reports, hand-authored
narrations of either solver fail to transfer, and only STaR on the model's own searches does.

\section{Training Configuration and Submission History}
\label{app:training}

I used two training stacks. The early ``lean'' lineage (HuggingFace \texttt{Trainer} $+$
PEFT) adapted only the projection modules. The later ``merge'' lineage (Unsloth $+$ a manual
loop), which produced the banked $0.85$ adapter, additionally adapts the attention
projections and the unembedding (\Cref{tab:trainconfig}). The trainable parameter count for
the projection-only set is $\approx\!880$M ($2.71\%$); the merge set is larger.

\begin{table}[htbp]
\centering
\caption{LoRA SFT configuration for the two training stacks. The banked $0.85$ adapter (v17)
uses the merge lineage. Cheap iteration used a managed LoRA service (Tinker) with the same
rank and learning rate. Its checkpoints require an SVD rank-$32$ conversion before
submission.}
\label{tab:trainconfig}
\small
\begin{tabular}{l l l}
\toprule
Setting & Lean lineage (v9--v14) & Merge lineage (v15--v19; banked v17) \\
\midrule
LoRA rank / $\alpha$ / dropout & $32$ / $16$ / $0.05$ & $32$ / $32$ / $0.0$ \\
target modules & in/out/up/down proj (4) & $+$ q/k/v/o proj, lm\_head (9) \\
learning rate & $2\times10^{-4}$ & $2\times10^{-4}$ \\
schedule & cosine, warmup $0.03$ & linear decay \\
optimizer & AdamW & AdamW $\beta(0.9,0.95),\ \varepsilon\,10^{-8}$, wd $0$ \\
effective batch & $16$ ($2\times8$) & $32$ (micro $4\times8$) \\
epochs & step-capped & $2$ (category-stratified) \\
max sequence length &, & $8192$ \\
precision & bf16 & LoRA fp32 / router fp32 / base bf16 \\
\bottomrule
\end{tabular}
\end{table}

\paragraph{Compute and the experiment ladder.} Final adapters train on a single RTX PRO 6000
(Blackwell, $94$--$96$\,GB). The first end-to-end run measured $\approx\!35$\,s/step. The
Blackwell GPU required a \texttt{ptxas} shim (locating the vendor binary, exporting
\texttt{TRITON\_PTXAS\_*}, and pinning a reported version) plus offline Mamba/causal-conv1d
kernels. I followed a three-tier cheapest-first ladder: (i) zero-cost base-model API probes
on the $30$B target and a $120$B reference; (ii) Tinker LoRA rounds (rank $32$, LR
$2\times10^{-4}$ fresh / $5\times10^{-6}$ resume, batch $32$, prompt tokens loss-masked) at a
few dollars per round and cents per held-out eval, under per-run dollar caps (GRPO runs
capped at \$1); (iii) the Kaggle from-base final, which avoids the SVD conversion tax that a
Tinker checkpoint incurs ($\approx\!0.025$ LB on my calibration).

\begin{table}[htbp]
\centering
\caption{Submission history. ``LB'' is the hidden-test score. The banked result is v17; v18
adds bit STaR within noise; v19 was parked. The null adapter (all $B=0$) reproduces the base
and certifies the pipeline.}
\label{tab:subs}
\small
\begin{tabular}{l c l l}
\toprule
Version & LB & Change & Note \\
\midrule
v9 / v10 & 0.52 & first end-to-end; easy-4 & \\
v12 & 0.57 & easy-4, improved cipher CoT & \\
v13 & 0.59 & $+$ early bit CoT & \\
v14 & 0.54 & per-bit tap CoT (regression) & lean \\
null adapter & 0.50 & $0$-step ($B{=}0$) & $\approx$ base $0.466$ \\
Tinker$\to$SVD & 0.79 & v16 corpus, converted & SVD tax \\
v15 & 0.82 & new grammars, warm-start & anchoring failure \\
v16 & 0.72 & compressed bit dialect & regression \\
\textbf{v17} & \textbf{0.85} & \textbf{from-base merge, $2e{-}4\times2$} & \textbf{banked} \\
v18 & 0.84 & $+$ bit STaR (additive) & within noise \\
v19 &, & $+$ eq/crypt bets & parked (no LB) \\
\bottomrule
\end{tabular}
\end{table}

\paragraph{Loss convergence vs.\ learning.} Every SFT run drove training loss to the floor
quickly and regardless of whether the procedure was learnable: the Kaggle merge runs log a
final loss $\approx\!0.003$, the Tinker warm-start reaches NLL $\approx\!0.001$, and the
cryptarithm rounds reach NLL $0.146$ by step $35$ and the floor by their second epoch. Yet
held-out cryptarithm accuracy stayed in $[0.01,0.07]$ across all fourteen rounds
(\Cref{tab:ledger}). Train loss is not a signal for this kind of learnability; only held-out
verification is. (Full per-step loss curves were not persisted; I recovered checkpoint NLLs
and the final-loss tail, so I report convergence points rather than curves.)

\begin{longtable}{l r r r r r l}
\caption{Full cheap-iteration run ledger (Tinker), the training log behind the experiment
ladder. ``Held-out acc'' is the run's target-category accuracy recovered from the eval oracle
(cryptarithm rounds report deduce accuracy. Bit rounds report overall bit accuracy);
$\ast$ is the multi-category weighted oracle. \$ is a list-price estimate. The pattern is the
paper in one table: cryptarithm rounds consume real compute and never move; bit climbs only
under STaR.}\label{tab:ledger}\\
\toprule
Run & Ex & Ep & Steps & min & \$$\sim$ & Held-out acc \\
\midrule
\endfirsthead
\multicolumn{7}{l}{\small\itshape (\Cref{tab:ledger}, continued)}\\
\toprule
Run & Ex & Ep & Steps & min & \$$\sim$ & Held-out acc \\
\midrule
\endhead
\midrule
\multicolumn{7}{r}{\small continued on next page}\\
\endfoot
\bottomrule
\endlastfoot
v16-warmstart   & 7987  & 1 & 249 & 22 & 9.0 & 0.82$^\ast$ \\
\addlinespace
crypt-boot-r1   & 1138  & 3 & 105 & 9  & 1.4 & 0.03 \\
crypt-boot-r2   & 3438  & 2 & 214 & 17 & 3.3 & 0.07 \\
crypt-boot-r3   & 3437  & 2 & 214 & 17 & 3.6 & 0.05 \\
crypt-boot-r5   & 3515  & 2 & 218 & 17 & 4.2 & 0.04 \\
crypt-boot-r6   & 3515  & 2 & 218 & 16 & 4.9 & 0.05 \\
crypt-boot-r7   & 3695  & 2 & 230 & 16 & 3.0 & 0.07 \\
crypt-boot-r8   & 3693  & 2 & 230 & 21 & 6.7 & 0.05 \\
crypt-twn-r9    & 3180  & 2 & 198 & 14 & 2.2 & 0.01 \\
crypt-twn-r10   & 2094  & 2 & 130 & 9  & 1.3 & 0.01 \\
crypt-twn-r11   & 2085  & 2 & 130 & 6  & 1.4 & 0.01 \\
crypt-honest-r1 & 1011  & 4 & 124 & 8  & 1.4 & 0.04 \\
crypt-honest-r2 & 719   & 4 & 88  & 6  & 1.3 & 0.03 \\
crypt-honest-r3 & 775   & 8 & 192 & 14 & 2.1 & 0.02 \\
crypt-grpo (RL) &, & 6\,it &, &, & 1.0 & 0.04 \\
\addlinespace
bit-global-r1   & 3302  & 2 & 206 & 15 & 4.1 & 0.52 \\
bit-global-r2   & 1800  & 2 & 112 & 9  & 2.3 & 0.15 \\
bit-r3          & 1673  & 2 & 104 & 8  & 3.6 & 0.53 \\
bit-dialect2    & 1101  & 2 & 68  & 5  & 3.8 & 0.05 \\
bit-star-r1     & 1984  & 2 & 124 & 9  & 6.4 & 0.66 \\
bit-star-r2     & 2163  & 2 & 134 & 11 & 7.0 & \textbf{0.68} \\
bit-3tap        & 4600  & 2 & 286 & 21 & 3.5 & 0.06 \\
bit-synth3      & 1836  & 2 & 114 & 9  & 6.1 & 0.60 \\
bit-peel        & 2000  & 3 & 186 & 13 & 2.3 & 0.09 \\
\addlinespace
v19-tinker      & 10662 & 1 & 333 & 74 & 8.9 & (merged) \\
\end{longtable}

\section{Tokenizer Analysis}
\label{app:tokenizer}

The tokenizer is a byte-level BPE: $131{,}072$ vocabulary entries, $269{,}443$ merges. \texttt{unk\_token} is \texttt{None} and byte-fallback is off (so coverage is via the
byte alphabet, and no \texttt{<unk>} is emitted in practice, though an \texttt{<unk>} symbol
occupies id $0$). There are $1{,}000$ added tokens, including the reasoning delimiters
\verb|<think>|/\verb|</think>| and $982$ reserved \verb|<SPECIAL_n>| slots.

\paragraph{ASCII-only is a real constraint.} Unicode logic/math glyphs fragment into rare
multi-byte pieces, which the model handles poorly even though no \texttt{<unk>} fires; ASCII
spellings are one or two common tokens (\Cref{tab:tok}). I therefore write \emph{all}
chain-of-thought in ASCII, operators as words (\texttt{XOR}/\texttt{OR}/\texttt{AND}/\texttt{NOT}),
\verb|->| for arrows, \verb|!=| for inequality, \texttt{ok}/\texttt{no} for $\checkmark$/$\neq$.
Digits tokenize one-per-digit, so an 8-bit string is exactly $8$ tokens, convenient for the
bit task.

\begin{table}[htbp]
\centering
\caption{Token counts for Unicode operators vs.\ their ASCII spellings under the competition
tokenizer (measured). The fragments are rare byte-pieces.}
\label{tab:tok}
\small
\begin{tabular}{l c | l c}
\toprule
Unicode & tokens & ASCII & tokens \\
\midrule
$\oplus$ (XOR) & 3 & \texttt{XOR} & 2 \\
$\checkmark$   & 3 & \texttt{OR}/\texttt{AND}/\texttt{NOT} & 1 \\
$\neq$         & 2 & \verb|->| & 1 \\
$\vee$ / $\wedge$ & 2 & \verb|!=| & 1 \\
$\neg$         & 2 & \texttt{ok}/\texttt{no} & 1 \\
$\to$ (rare)   & 1 & \texttt{XNOR} & 3 \\
\bottomrule
\end{tabular}
\end{table}

\paragraph{Lengths (measured on the actual data).} Prompts are short and CoT terser still
(\Cref{tab:toklen}), both far under the $7680$-token budget. The gap between the
$\leq\!274$-token synthetic CoT I \emph{train} on and the $\approx\!1$k--$3.8$k tokens the
model \emph{generates} at inference (\Cref{tab:tokens}) is the base's native verbosity
reasserting itself, and is part of why authored brevity does not survive to inference.

\begin{table}[htbp]
\centering
\caption{Prompt and synthetic-CoT token lengths by category (measured with the competition
tokenizer over all training rows). Cryptarithm CoT was rendered per round (\Cref{sec:res-crypt}),
not in this synthetic corpus.}
\label{tab:toklen}
\small
\begin{tabular}{l r r}
\toprule
Category & prompt (med / max) & synth CoT (med / max) \\
\midrule
numeral          & 61 / 72   & 99 / 154 \\
unit\_conversion & 87 / 102  & 181 / 184 \\
gravity          & 128 / 152 & 150 / 157 \\
cipher           & 118 / 170 & 168 / 222 \\
bit\_manipulation& 240 / 259 & 221 / 274 \\
equation\_numeric& 75--80 / 96 & 188 / 213 \\
cryptarithm      & 62--64 / 79 &, (per round) \\
\bottomrule
\end{tabular}
\end{table}

\paragraph{Budget.} The $7680$-token budget (with $\texttt{max\_model\_len}=8192$) is
comfortable for most categories. Free-run generation medians are $\approx\!1$k (numeral) to
$\approx\!3.8$k (bit). It binds only on rendered cryptarithm \emph{search} traces, which loop
to the cap, one reason STaR's shorter self-traces, which cut bit truncation from $18.6\%$ to
$0.2\%$, helped.

\paragraph{Per-token difficulty audit.} Logging per-token cross-entropy over the training
CoT ($6{,}755$ token types) gives a vocabulary-discipline tool (\Cref{tab:tokce}). Two things
stand out. First, the count-weighted mean CE ranks the categories: the lookup/fit tasks sit
at $2.1$--$3.2$ while cryptarithm and cipher sit at $5.9$--$6.3$, the harder categories are
written in tokens the base finds more surprising (rare themed cipher words, cryptarithm
glyphs, enumeration). Second, even the boolean \emph{operator} words I adopted to stay ASCII
carry real surprise (\texttt{->} CE $6.9$ over $333$k uses, \texttt{OR} $6.5$, \texttt{NOT}
$7.3$, \texttt{AND} $8.7$, \texttt{majority} $10.4$), reinforcing the native-voice finding:
the CoT vocabulary is itself somewhat out-of-distribution for the base. The audit flags a
$1{,}253$-token ``blacklist'' (CE $\geq 7$, seen $\geq 100\times$; $771$ of them cipher
vocabulary) that a linter then keeps out of authored traces. (High token CE is a rarity, not
an accuracy, signal: cipher has the highest token CE yet $\approx\!0.99$ accuracy, because its
rare tokens are answer-bearing vocabulary words.)

\begin{table}[htbp]
\centering
\caption{Count-weighted mean per-token cross-entropy of the training CoT, by category (token
audit. Lower $=$ the base finds the trace tokens less surprising). The lookup/fit tasks are
cheapest. Cryptarithm and cipher are written in the most surprising tokens.}
\label{tab:tokce}
\small
\begin{tabular}{l r | l r}
\toprule
Category & mean CE & Category & mean CE \\
\midrule
unit\_conversion & 2.09 & equation\_num.\ deduce & 4.97 \\
gravity          & 2.27 & equation\_num.\ guess  & 5.19 \\
numeral          & 3.21 & cryptarithm deduce     & 5.85 \\
bit\_manipulation& 4.15 & cryptarithm guess      & 5.98 \\
                 &      & cipher                 & 6.34 \\
\bottomrule
\end{tabular}
\end{table}

\section{Per-Sub-Category Results and Failure Statistics}
\label{app:failstats}

\Cref{tab:persubcat} gives accuracy for the base model, a $0.86$-class reference adapter
(measured by re-running it over all $9{,}500$ training rows), and my banked adapter (on a
held-out ``unseen'' split, the trustworthy generalization number). \Cref{tab:tokens} gives
the length and failure-type breakdown that grounds \Cref{sec:anatomy}: for the shipped
adapter, failures are wrong-answer rather than truncation in every category except the
cryptarithm search renders.

\begin{table}[htbp]
\centering
\caption{Per-sub-category accuracy. The reference adapter is measured by direct forensic
re-run. Small-$n$ ``unseen'' cells for my adapter (bit, eq-deduce) are superseded by
disjoint evaluations where noted in the text (bit $=0.678$ on a $500$-row set; eq-deduce
$\approx\!0.85$).}
\label{tab:persubcat}
\small
\begin{tabular}{l c c c c}
\toprule
Sub-category & Test wt. & Base & Ref.\ adapter ($\approx$0.86) & Ours (unseen) \\
\midrule
numeral            & 16.6\% & 0.959 & 1.000 & $\approx$1.00 \\
unit\_conversion   & 16.8\% & 0.745 & 0.997 & $\approx$1.00 \\
gravity            & 16.8\% & 0.585 & 0.997 & $\approx$1.00 \\
cipher             & 16.6\% & 0.306 & 0.992 & $\approx$0.99 \\
bit\_manipulation  & 16.9\% & 0.080 & 0.767 & \textbf{0.678} \\
eq\_num.\ deduce   & 6.3\%  & 0.287 & 0.906 & $\approx$0.85 \\
eq\_num.\ guess    & 1.4\%  & 0.037 & 0.154 & 0.19 \\
cryptarithm deduce & 6.9\%  & 0.003 & 0.084 & 0.01--0.03 \\
cryptarithm guess  & 1.7\%  & 0.000 & 0.067 & 0.02 \\
\midrule
\multicolumn{2}{l}{overall} & 0.466 & 0.861 & \textbf{0.85} \\
\bottomrule
\end{tabular}
\end{table}

\begin{table}[htbp]
\centering
\caption{Length and failure type for the banked adapter (100 rows/category). ``Wrong-finished''
means the model emitted an answer within budget and it was wrong; ``trunc'' means it hit the
$7680$-token cap. Only the cryptarithm \emph{rendered-search} corpora (not shown; a different
training-data choice) truncate heavily ($64$--$72\%$).}
\label{tab:tokens}
\small
\begin{tabular}{l r r l}
\toprule
Category & median tokens & trunc.\ \% & dominant failure \\
\midrule
numeral            & 955  & 0   &, (100\% correct) \\
cipher             & 2153 & 0   & unseen-letter rows \\
unit\_conversion   & 2222 & 0   &, \\
gravity            & 3175 & 0   &, \\
bit\_manipulation  & 3783 & 0   & 17\% wrong-finished (3-tap) \\
eq\_num.\ deduce   & 600  & 0--1 & arithmetic slips \\
eq\_num.\ guess    & 559  & 0--1 & 79\% wrong-finished (unguessable op) \\
cryptarithm deduce & 486  & 1--3 & 96\% wrong-finished (verdict-as-token) \\
cryptarithm guess  & 498  & 0--8 & 90\% wrong-finished \\
\bottomrule
\end{tabular}
\end{table}

\section{Leaderboard Trajectory}
\label{app:lbtraj}
\begin{figure}[htbp]
\centering
\begin{tikzpicture}
\begin{axis}[
  width=0.92\textwidth, height=5.4cm,
  ybar, bar width=11pt,
  ymin=0.45, ymax=0.95,
  ylabel={Hidden-test accuracy},
  symbolic x coords={base, v9, v10, v12, v13, v14, v15, v16, v17, v18},
  xtick=data, x tick label style={font=\footnotesize},
  ytick={0.5,0.6,0.7,0.8,0.9},
  nodes near coords, nodes near coords style={font=\tiny,/pgf/number format/fixed,/pgf/number format/precision=2},
  enlarge x limits=0.06,
]
\addplot[fill=c0,draw=c0] coordinates {
  (base,0.50) (v9,0.52) (v10,0.52) (v12,0.57) (v13,0.59) (v14,0.54)
  (v15,0.82) (v16,0.72) (v17,0.85) (v18,0.84)};
\draw[c1,thick,dashed] (axis cs:base,0.92) -- (axis cs:v18,0.92)
  node[above left,font=\footnotesize,c1] {leader (NullSira) 0.92};
\draw[c2,thick,dotted] (axis cs:base,0.87) -- (axis cs:v18,0.87)
  node[below left,font=\footnotesize,c2,yshift=-1pt] {top-decile $\approx$0.87};
\end{axis}
\end{tikzpicture}
\caption{\textbf{Leaderboard trajectory.} v15$\to$v16 is a regression from a warm-start that
failed to enter the new grammars at greedy decoding; v17 ($0.85$) is the first reproducible
from-base bank. V18 adds bit-manipulation STaR but does not move the total, the headroom that
remains is cryptarithm.}
\label{fig:lb}
\end{figure}
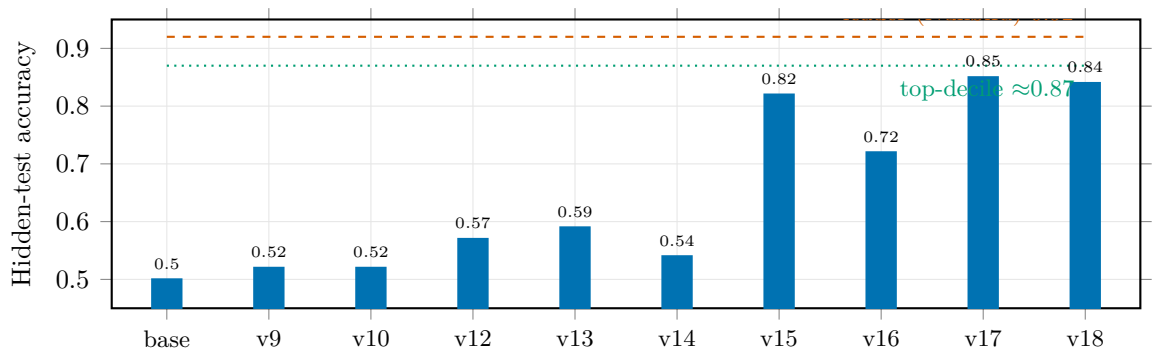

For context, my submission history (\Cref{fig:lb}). Gains come from making more categories
learnable, not from polishing a single model.

\bibliographystyle{plainnat}
\bibliography{references}

\end{document}